\documentclass[runningheads]{llncs}
\usepackage{graphicx}

\usepackage{orcidlink}

\usepackage{tikz}
\usepackage{comment}
\usepackage{amsmath,amssymb} 
\usepackage{color}
\usepackage{appendix}
\usepackage{booktabs} 
\usepackage{hyperref}
\usepackage{subcaption}
\usepackage[ruled,vlined, linesnumbered]{algorithm2e}
\SetKwInput{KwInput}{Input}                
\SetKwInput{KwOutput}{Output}              
\SetKwComment{Comment}{$\triangleright$\ }{}
\usepackage[switch]{lineno}
\usepackage{makecell}
\usepackage[symbol]{footmisc}

\usepackage{wrapfig}
\usepackage[accsupp]{axessibility}  

\usepackage[textsize=small]{todonotes}

\newcommand*\samethanks[1][\value{footnote}]{\footnotemark[#1]}

\begin{document}
\pagestyle{headings}
\mainmatter
\def\ECCVSubNumber{6011}  

\title{Learning Where To Look -- \\Generative NAS is Surprisingly Efficient}

\titlerunning{Learning Where To Look -- Generative NAS is Surprisingly Efficient}
%
\author{
  Jovita Lukasik\textsuperscript{\rm 1,2}\thanks{Authors contributed equally.}\orcidlink{0000-0003-4243-9188} \and Steffen Jung\textsuperscript{\rm 2}\samethanks\orcidlink{0000-0001-8021-791X} \and Margret Keuper\textsuperscript{\rm 2,3}\orcidlink{0000-0002-8437-7993}}
\authorrunning{J. Lukasik, S. Jung, M. Keuper}
%
\institute{University of Mannheim\and
Max Planck Institute for Informatics, Saarland Informatics Campus
\and
University of Siegen \\
}
\maketitle

\begin{abstract}
The efficient, automated search for well-performing neural architectures (NAS) has drawn increasing attention in the recent past. 
Thereby, the predominant research objective is to reduce the necessity of costly evaluations of neural architectures while efficiently exploring large search spaces. 
To this aim, surrogate models embed architectures in a latent space and predict their performance, while generative models for neural architectures enable optimization-based search within the latent space the generator draws from. Both, surrogate and generative models, have the aim of facilitating query-efficient search in a well-structured latent space. In this paper, we further improve the trade-off between query-efficiency and promising architecture generation by leveraging advantages from both, efficient surrogate models and generative design. To this end, we propose a generative model, paired with a surrogate predictor, that iteratively learns to generate samples from increasingly promising latent subspaces.  
This approach leads to very effective and efficient architecture search, while keeping the query amount low. 
In addition, our approach allows in a straightforward manner to jointly optimize for multiple objectives such as accuracy and hardware latency. 
We show the benefit of this approach not only w.r.t. the optimization of architectures for highest classification accuracy but also in the context of hardware constraints and outperform state-of-the-art methods on several NAS benchmarks for single and multiple objectives. We also achieve state-of-the-art performance on ImageNet. 
The code is available at \url{https://github.com/jovitalukasik/AG-Net}.

\end{abstract}

\section{Introduction}

\begin{figure}[t]

    \centering
    \begin{tabular}{c}
        \includegraphics[height=2.7cm]{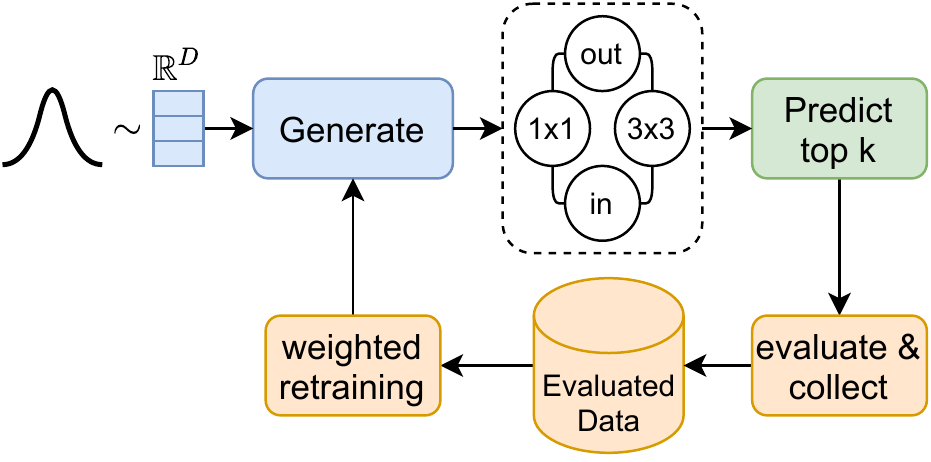} 
        \includegraphics[height=2.7cm]{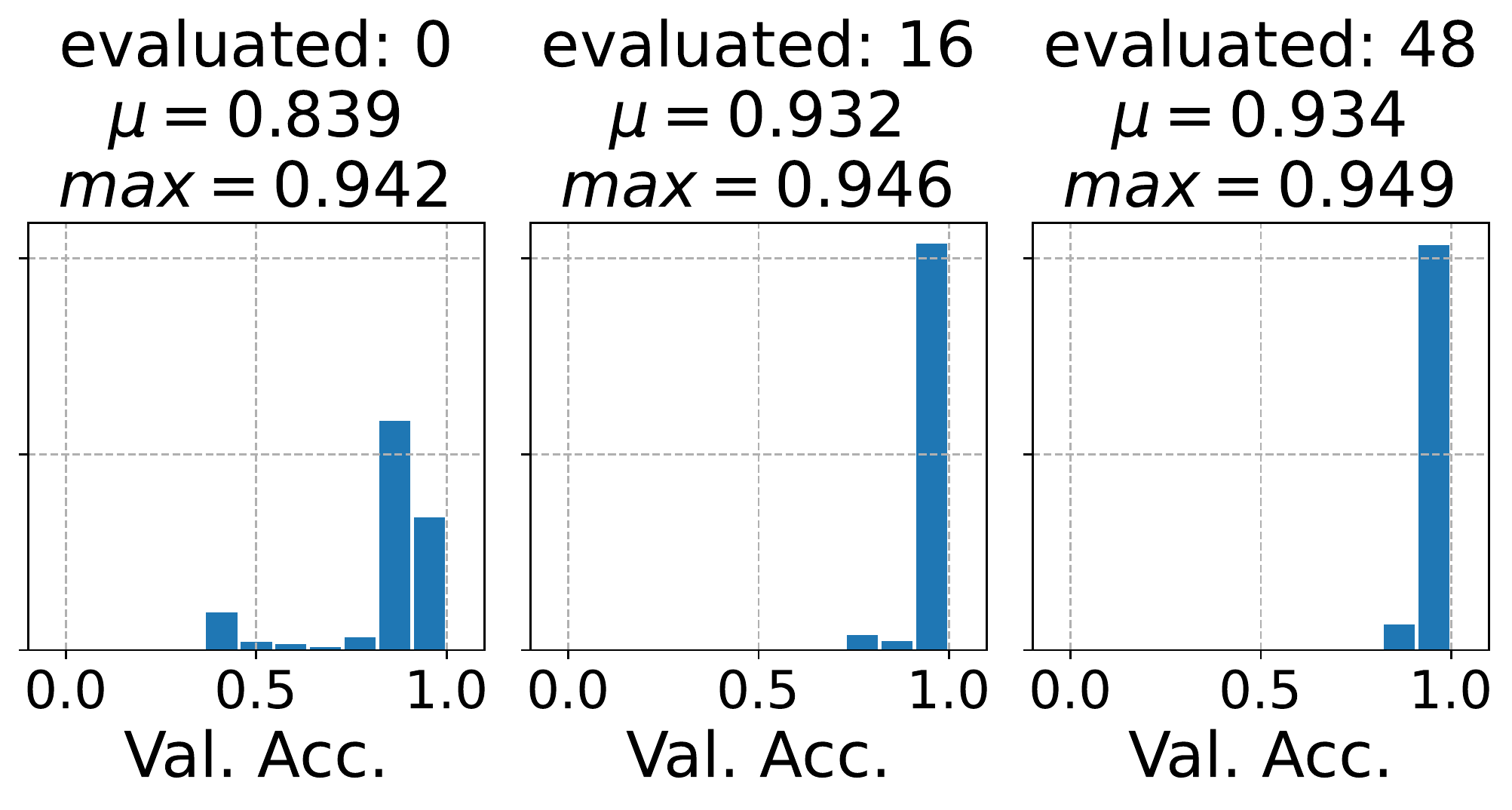}
    \end{tabular}
    \caption{
        (\textbf{left}) Our search method generates architectures from points in an architecture representation space that is iteratively optimized.
        (\textbf{right}) The architecture representation space is biased towards better-performing architectures with each search iteration.
        After only $48$ evaluated architectures, our generator produces state-of-the-art performing architectures on NAS-Bench-101.
    }
    \label{fig:teaser}

\end{figure}

The first image classification network \cite{2012AlexNet} applied to the large-scale visual recognition challenge ImageNet \cite{2009ImageNet} achieved unprecedented results. Since then, the main driver of improvement on this challenge are new architecture designs \cite{2014VGG,2015GoogleNet}, \cite{2016Inception,2016ResNet} that, ultimately, lead to architectures surpassing human performance \cite{2015ReLU}.
Since manual architecture design requires good intuition and a huge amount of trial-and-error, the automated approach of neural architecture search (NAS) receives growing interest \cite{2017EvolutionaryNAS,2018LearningNAS,2019NB101,2020NB201,2020NBNLP,2021HWNNB}. Well-performing architectures can be found by applying common search practices like random search \cite{2012RandomNAS}, evolutionary search \cite{2017EvolutionaryNAS,2019EvolutionaryNAS}, Bayesian optimization (BO) \cite{2018BONAS,2020BONAS,2021BANANAS}, or local search \cite{2020LocalSearchNAS} on discrete architecture search spaces, such as NAS-Bench-101, NAS-Bench-201, DARTS and NAS-Bench-NLP  \cite{2019NB101,2020NB201,2018DARTS,2020NBNLP}. However, these methods are inefficient because they require to evaluate thousands of architectures, resulting in impracticable search times. Recent approaches avoid this problem of immense computation costs by either training surrogate models to approximate the performance of an architecture \cite{2018DARTS,2019ProxyNAS} or by generating architectures based on learned architecture representation spaces \cite{2019VAENAS,2021SVGe}.
Both methods aim to improve the query efficiency, which is crucial in NAS, since every query implies a full training and evaluation of the neural architecture on the underlying target dataset.

This trade-off between query efficiency and resulting high-scoring architectures is an active research field. Yet, no attempts were made so far to leverage the advantages of both search paradigms. Therefore, we propose a model that incorporates the focus of promising architectures already in the architecture generation process by optimizing the latent space \textit{directly}: We let the generator learn in which areas of the data distribution to look for promising architectures. This way, we reduce the query amount even further, resulting in a query efficient and very effective NAS method. 
Our proposed method is inspired by a latent space optimization (LSO) technique \cite{2020Reweighting}, originally used in the context of variational autoencoders \cite{2014VAE} to optimize generated images or arithmetic expressions using BO. We adapt this concept to NAS and pair it with an architecture performance predictor in an end-to-end learning setting, so that it allows us to iteratively reshape the architecture representation space. Thereby, we promote desired properties of generated architectures in a highly query-efficient way, i.e.~by learning expert generators for promising architectures.
Since we couple the generation process with a surrogate model to predict desired properties such as high accuracy or low latency of generated architectures, there is no need in our method for BO in the generated latent space, making our method even more efficient. 

In practice, we pretrain, on a target space of neural architectures, a GNN-based generator network, which does not rely on any architecture evaluation and is therefore fast and query-free. 
The generator is trained in a novel generative setting that directly compares generated architectures to randomly sampled architectures using a reconstruction loss without the need of a discriminator network as in generative adversarial networks (GANs) \cite{2014GAN} or an encoder as in variational autoencoders (VAEs) \cite{2014VAE}.
We use an MLP as a surrogate to rank performances and hardware properties of generated architectures. In contrast, previous generative methods either rely on training and evaluating supernets \cite{2021SGNAS}, which are expensive to train and dataset specific, or pretrain a latent space and search within this space directly using BO \cite{2019VAENAS,2020Arch2vec,2021SVGe}, reinforcement learning (RL) \cite{2021GANAS} or gradient based methods \cite{2018NAO}. These methods incorporate either GANs, which can be hard to train or VAEs, which are biased by the regularization, whereas our plain generative model is easy to train. 
In addition we enable backpropagation from the performance predictor to the generator. Thereby, the generator can efficiently learn which part of the architecture search space is promising with only few evaluated architectures.

By extensive experiments on common NAS benchmarks \cite{2019NB101,2020NB201,2020NB301,2020NBNLP,2021HWNNB} as well as ImageNet \cite{2009ImageNet}, we show that our method is effective and sample-efficient. It reinforces the generator network to produce architectures with improving validation accuracy (see \autoref{fig:teaser}), as well as in improving on hardware-dependent latency constraints (see \autoref{fig:feasibility}) while keeping the number of architecture evaluations small.
In summary, we make the following contributions:
\begin{itemize}
    \item We propose a simple model that learns to focus on promising regions of the architecture space. It can thus learn to generate high-scoring architectures from only a few queries. 
    \item We learn architecture representation spaces via a \textit{novel generative design} that is able to generate architectures stochastically 
    while being trained with a simple reconstruction loss.
    Unlike VAEs \cite{2014VAE} or GANs \cite{2014GAN}, no encoder network nor discriminator network is necessary.
    \item Our model allows sample-efficient search and achieves state-of-the-art results on several NAS benchmarks as well as on ImageNet. It allows joint optimization w.r.t. hardware properties in a straightforward way.
\end{itemize}

\section{Related Work}
\paragraph{Neural Architecture Search}
Intrinsically, Neural Architecture Search (NAS) is a discrete optimization problem seeking the optimal configuration of operations (such as convolutions, poolings and skip connections) in a constrained \emph{search space} of computational graphs.
To enable benchmarking within the NAS community, different search spaces have been proposed. The tabular benchmarks NAS-Bench-101 \cite{2019NB101} and NAS-Bench-201 \cite{2020NB201} provide both an exhaustive covering of metrics and performances. NAS-bench-NLP \cite{2020NBNLP} provides a search space for natural language processing. 
In addition to these tabular benchmarks NAS-Bench-301 \cite{2020NB301} provides a surrogate benchmark, which allows for fast evaluation of NAS methods on the DARTS \cite{2018DARTS} search space by querying the validation accuracy. 
NAS-Bench-x11 \cite{2021NBX11} is another  surrogate benchmark. It outputs full training information for each architecture in all four mentioned benchmarks. NAS-Bench-Suite \cite{NBSuite} facilitates reproducible search on these NAS benchmarks.

Early NAS approaches are based on discrete encodings of search spaces, such as in the form of adjacency matrices, and can be distinguished by their \emph{search strategy}.
Examples are random search \cite{2012RandomNAS,2019RS}, reinforcement learning (RL) \cite{2017ReinforcementNAS,2018ReinforcementNAS}, evolutionary methods \cite{2017EvolutionaryNAS,2019EvolutionaryNAS}, local search \cite{2020LocalSearchNAS}, and Bayesian optimization (BO) \cite{2018BONAS,2020BONAS}.
Recent NAS methods shift from discrete optimization to faster weight-sharing approaches, resulting in differentiable optimization methods \cite{2018ParameterSharingNAS,2018DARTS,2018ONESHOT,2019ProxyNAS,2019SNAS,2020RobustDarts}.
Several approaches map the discrete search space into a continuous architecture representation space \cite{2018NAO,2019VAENAS,2020Arch2vec,2021SVGe} and search or optimize within this space using for example BO (e.g.~\cite{2020Arch2vec}) or gradient-based point operation \cite{2018NAO}. 
In this paper, we also learn continuous architecture representation spaces.
However, in contrast to former works, we propose to optimize the representation space, instead of performing point optimization within a fixed space such as e.g.~\cite{2018NAO}. 
A survey of different strategies can be found in \cite{2019NASSurvey}.

All NAS approaches are dependent on \emph{performance estimation} of intermediate architectures.
To avoid the computation heavy training and evaluation of queries on the target dataset, methods to approximate the performance have been explored \cite{2021HowPP}.
Common approaches include neural predictors that take path encodings \cite{2021BANANAS} or graph embeddings learned by GNNs \cite{2019NASPredictor,2020NP} as input.
WeakNAS \cite{2021WeakNAS} proposes to progressively evaluate the search space towards finding high-performing architectures using a set of weak predictors. 
In our method, we integrate a weak expert predictor with a generator to yield an efficient interplay between predicting for high-performing architectures and generating them.   

\noindent\textit{Graph Generative Models}
Most graph generation models in NAS employ variational autoencoders (VAE) \cite{2014VAE}. 
\cite{2018NAO} uses an LSTM-based VAE, coupled with performance prediction for gradient-based architecture optimization. Note that \cite{2018NAO} optimizes the latent point in a fixed latent space while our approach optimizes the latent space itself. 
\cite{2019VAENAS} use GNNs with asynchronous message-passing to train a VAE for BO.
\cite{2021SGNAS} combines a generator with a supernet and searches for neural architectures for different device information.
\cite{2020Arch2vec} facilitates \cite{2019GIN} with an MLP decoder.
\cite{2021SVGe} proposes smooth variational graph embeddings (SVGe) 
using two-sided GNNs 
to capture the 
information flow within a neural architecture.

Our proposed model's generator is inspired by SVGe with the aim to inherit its flexible applicability to various search spaces. Yet, similar to \cite{2020Arch2vec}, due to the intrinsic discretization and training setting, SVGe does not allow for backpropagation. 
Recently, \cite{2021GANAS} facilitates GNNs in a GAN \cite{2014GAN} setting, where the backpropagation issue is circumvented using reinforcement learning. 
In contrast, our proposed GNN generator circumvents the intermediate architecture discretization and can therefore be trained by a single reconstruction loss using backpropagation.
Its iterative optimization is inspired by \cite{2020Reweighting}, who proposes to use a VAE with weighted retraining w.r.t.~a target function to adapt the latent space for the optimization of images and arithmetic functions using BO. Our model transfers the idea of weighted retraining to NAS. It uses our plain generator and improves sample efficiency by employing a differentiable surrogate model on the target function such that, in contrast to \cite{2020Reweighting}, no further black-box optimization step is needed. Next, we describe the proposed generator network.


\begin{figure}[t!]

    \centering
    \includegraphics[height=3cm]{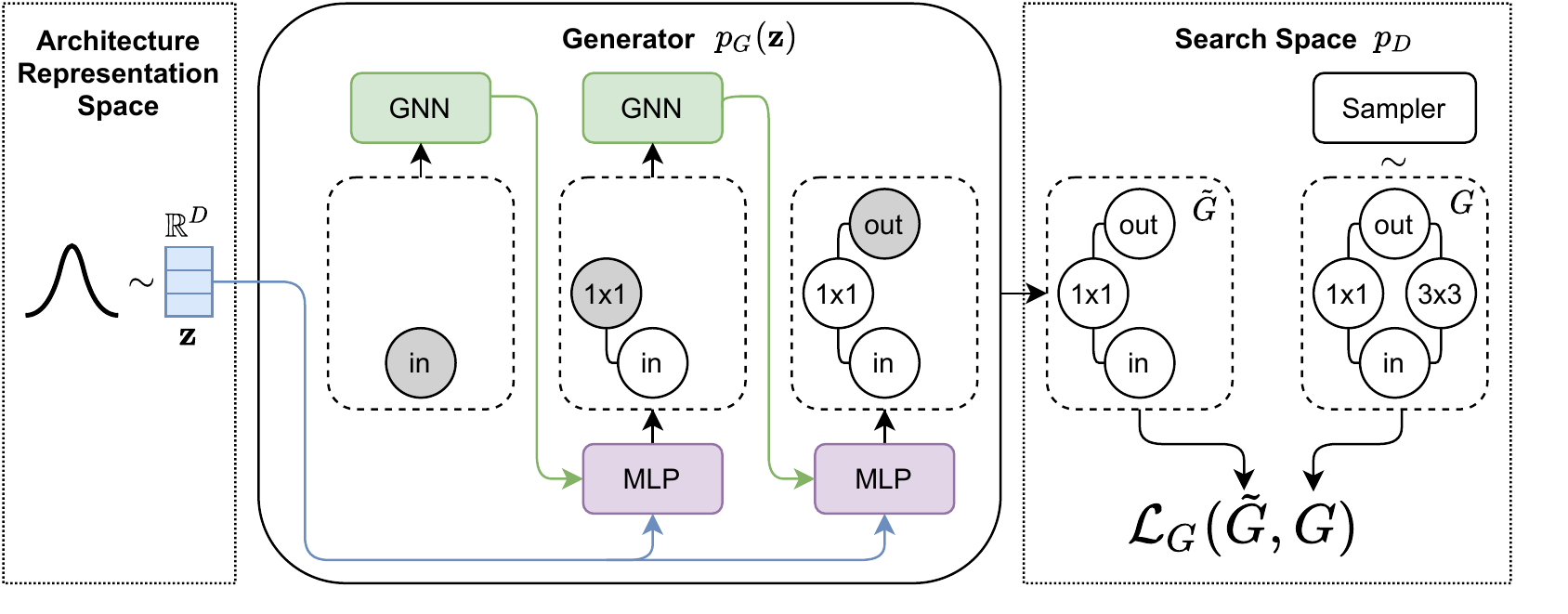}
    \caption{Representation of the training procedure for our generator in AG-Net. The input is a randomly sampled latent vector $\mathbf{z} \in \mathbb{R}^d$. First, the input node is generated, initialized and input to a GNN to generate a partial graph representation. The learning process iteratively generates node scores and edge scores using $\mathbf{z}$ and the partial graph representation until the output node is generated. The target for this generated graph is a randomly sampled architecture.    
    }
    \label{fig:generator-training}
\end{figure}

\section{Architecture Generative Model}
\noindent \textit{Preliminaries}
We aim to generate neural networks represented as directed acyclic graphs (DAG).
This representation is in line with the cell-based architecture search spaces commonly used as tabular benchmarks \cite{2019NB101,2020NB201}.
Each cell is a DAG $G=(V,E)$, with nodes $v \in V$ and edges $e \in E$.
The graph representations differ between the various benchmarks in terms of their labeling of operations.
For example in NAS-Bench-101 \cite{2019NB101} each node is associated with an operation, whereas in NAS-Bench-201 \cite{2020NB201} each edge is associated with an operation.

\noindent \textbf{Generative Network}
Commonly used graph generative networks are based on variational  autoencoders (VAE)~\cite{2014VAE}.
In contrast, our proposed network is a \emph{purely generative} network, $p_G$ (see \autoref{fig:generator-training}).
To generate valid graphs, we build our model similar to the graph decoder from the VAE approach SVGe~\cite{2021SVGe}. The generator takes a randomly sampled variable $\textbf{z} \sim \mathcal{N}(0,1)$ as input and reconstructs a randomly sampled graph from the cell-based search space. The model iteratively builds the graph: it starts with generating the input node $v_\textrm{0}$, followed by adding subsequent nodes $v_\textrm{i}$ and their labels and connecting them with edges $e_{(j,i)}, j<i$, until the end node $v_\textrm{T}$ with the label \textit{output} is generated. Additionally, we want to learn a surrogate for performance prediction on the generated data and allow for end-to-end training of both. 
To allow for backpropagation, we need to adapt several details of the generator model. We initialize the node-attributes for each node by one-hot encoded vectors, which are initialized during training using a 2-layer MLP to replace the learnable look-up table proposed in SVGe. The output of our generator is a vector graph representation consisting of a concatenation of generated node scores and edge scores. It is important to note that the iterative generation process is independent of the ground truth data, which are only used as a target for the reconstruction loss. Note that the end-to-end trainability of the proposed generator is a prerequisite for our model: It allows to pair the generator with a learnable performance predictor such that information on the expected architectures' accuracy can be learned by the generator. This enables a stronger coupling with the predictor's target for the generation process and higher query efficiency (see \autoref{sec:ablation_studies}). In contrast, previous models such as \cite{2021SGNAS,2021SVGe,2020Arch2vec} are not fully differentiable and do not allow such optimization.
Our generative model is pretrained on the task of reconstructing neural architectures, where for each randomly drawn latent space sample, we evaluate the reconstruction loss to a randomly drawn architecture.
This simple procedure is facilitated by the heavily constrained search spaces of neural architectures, making it easy for the model to learn to generate valid architectures without being supported by a discriminator model as in generative adversarial networks (GANs) \cite{2014GAN}. An evaluation of the generation ability of our model and implementation details are provided in the supp.~mat.~\autoref{supp:sec_generator_ability}.

\noindent \textbf{Performance Predictor}
This generative model is coupled with a simple surrogate model, a 4-layer MLP with ReLU non-linearities, for target predictions $C$. These targets can be validation or test accuracy of the generated graph, or the latency with respect to a certain hardware. 
For comparison, we also include a tree-based method, XGBoost (XGB) \cite{XGB} as an alternative prediction model. XGB\cite{XGB} is used as a surrogate model in NAS-Bench-301 \cite{2020NB301} and shows high prediction abilities. The input to XGB is the vector representation of the architectures. Since this method is non-differentiable, we additionally include a gradient estimation for rank-based metrics \cite{RankbasedGradients}. This way, we are able to include gradient information to the generator. Yet, it is important to note, that this approach is not fully differentiable. This comparison will allow us to measure the trade-off between using supposedly stronger predictors over the capability to allow for full end-to-end learning.

\noindent \textbf{Training Objectives}
The generative model $p_G$ learns to reconstruct a randomly sampled architecture $G$ from search space $p_D$ given a randomly sampled latent vector $\textbf{z} \sim \mathcal{N}(0,1)$.
The objective function for this generation process can be formulated as the sum of node-level loss ${\mathcal{L}}_V$ and edge-level loss ${\mathcal{L}}_E$:
\begin{equation}\label{eq:generator_loss}
{\mathcal{L}}_G(\tilde{G},G) = {\mathcal{L}}_{V} + {\mathcal{L}}_E;
\;\tilde{G}\sim p_G(\textbf{z});\;G\sim p_D, \
\end{equation}
where ${\mathcal{L}}_V$ is the Cross-Entropy loss between the predicted and the ground truth nodes and ${\mathcal{L}}_E$ is the Binary-Cross Entropy loss between the predicted and ground truth edges of the generated graph $\tilde{G}$.
This training step is \emph{completely unsupervised}.
\autoref{fig:generator-training} presents an overview of the training process.
To include the training of the surrogate model, the objective function is reformulated to:
\begin{equation}\label{eq:search_loss}
{\mathcal{L}}(\tilde{G},G) = (1-\alpha){\mathcal{L}}_G(\tilde{G}, G) + \alpha {\mathcal{L}_C}(\tilde{G},G),
\end{equation}
where $\alpha$ is a hyperparameter to trade-off generator loss $\mathcal{L}_G$ and prediction loss $\mathcal{L}_C$ for the prediction targets $C$ of graph $G$.
We set the predictor loss as an MSE. Furthermore, each loss is optimized using mini-batch gradient descent.

\noindent \textbf{Generative Latent Space Optimization (LSO)}
To facilitate the generation process, we optimize the architecture representation space via weighted retraining \cite{2020Reweighting}, resulting in a sample efficient search algorithm.
The intuition of this approach is to place more probability mass on high-scoring latent points, (e.g. high performing or low latency architectures) and less mass on low-scoring points.
Thus, this strategy does not discard low-scoring architectures completely, which would be inadequate for proper learning.
The generative model is therefore trained on a data distribution that systematically increases the probability of high-scoring latent points.
This can be done by simply assigning a weight $w_i$ to each data point $G_i \sim p_D$
, indicating its likelihood to occur during batch-wise training.
In addition, the training objective is weighted via a weighted empirical mean
$\sum_{G_i \sim p_D} w_i~\mathcal{L}$ 
for each data point.
As for the weights itself,  \cite{2020Reweighting} proposed a rank-based weight function
\begin{align}\label{eq:weights}
\begin{split}
&w(G;p_D,k) \propto \frac{1}{kN + \textrm{rank}_{f,p_D}(G)}\\
&\textrm{rank}_{f,p_D}(x) = \vert\{G_i : f(G_i) > f(G), G_i \sim p_D\} \vert,
    \end{split}
\end{align}
where $f(\cdot)$ is the evaluation function of the architecture $G_i$; for NAS-Bench-101 \cite{2019NB101} and NAS-Bench-201 \cite{2020NB201} it is the tabular benchmark entry, for NAS-Bench-301 \cite{2020NB301} and NAS-Bench-NLP \cite{2020NBNLP} it is the surrogate benchmark prediction.
Similar to \cite{2020Reweighting}, we set $k = 10e-3$.
The retraining procedure itself then consists of finetuning the pretrained generative model coupled with the surrogate model, where loss functions and data points are both weighted by $w(G;p_D,k)$.


\section{Experiments}\label{sec:Experiments}
We evaluate the proposed simple architecture generative network (AG-Net) on the two commonly used tabular benchmarks NAS-Bench-101 \cite{2019NB101} and NAS-Bench-201 \cite{2020NB201}, the surrogate benchmarks NAS-Bench-301 \cite{2020NB301} evaluated on the DARTS search space \cite{2018DARTS}, NAS-Bench-NLP \cite{2020NBNLP} and the first hardware device induced benchmark \cite{2021HWNNB}. Additionally we perform experiments on the ImageNet \cite{2009ImageNet} classification task and show state-of-the-art performance on the DARTS search space. In our experiments in \autoref{sec:hwbench-experiments} for the Hardware-Aware Benchmark we consider the latency information on the NAS-Bench-201 search space. 
Details about all hyperparameters are given in the supp. mat. \autoref{sec:HP}.


\subsection{Experiments on Tabular Benchmarks}\label{sec:tabluar_benchmark_experiments}
\noindent \textit{NAS-Bench-101}
For our experiments on NAS-Bench-101, we first pretrain our generator for generating valid graphs on the NAS-Bench-101 search space.
This step does not require information about the performance of architectures and is therefore inexpensive.
The pretrained generator is then used for all experiments on NAS-Bench-101.
Our NAS algorithm is initialized by randomly sampling $16$ architectures from the search space, which are then weighted by the weighting function $\mathcal{W} = {w(G)}_{G \sim p_D}$. 
Then, latent space optimized architecture search is performed by iteratively retraining the generator coupled with the MLP surrogate model 
for $15$ epochs and generating $100$ architectures of which the top $16$ (according to their accuracy prediction) are evaluated and added to the training data.
This step is repeated until the desired number of queries is reached.
When generating architectures
, we sample from a grid, containing the $99 \%$-quantiles from $\mathcal{N}(0,1)$ uniformly distributed.
This way, we sample more distributed latent variables for better latent space coverage.  
We compare our method to the VAE-based search method Arch2vec \cite{2020Arch2vec} and predictor based model WeakNAS \cite{2021WeakNAS}, as well as state-of-the-art methods, such as NAO \cite{2018NAO}\footnote[3]{We reran this experiment using the implementation from \cite{2021HowPP}.},  random search \cite{2019RS}, local search \cite{2020LocalSearchNAS}, Bayesian optimization \cite{2015DNGO}, regularized evolution \cite{2019EvolutionaryNAS} and BANANAS \cite{2021BANANAS}\footnote[2]{We reran these experiments using the official implementation from 
\cite{2020Study,2021BANANAS,2020LocalSearchNAS}, with the same initial training data and amount of top k architectures as for AG-Net.}.
Additionally, we compare the proposed AG-Net to the model using an XGBoost Predictor (see \autoref{sup:implementation-details}).
The results of this comparison are listed in \autoref{tab:NB101_Search}.
Here, we report the mean over 10 runs. Results including the standard deviation can be found in the supp.~mat.
Note, we search for the architecture with the best validation accuracy and report the corresponding test accuracy.
Furthermore, we plot the search progress in \autoref{fig:NB101_search} (bottom left).
As we can see, our model AG-Net improves over all state-of-the-art methods, not only at the last query of $300$ data points, reaching a top 1 test accuracy of $94.2 \%$, but is also almost any time better during the search process.

A direct comparison to the recently proposed GANAS \cite{2021GANAS} on NAS-Bench-101 is difficult, since GANAS searches on NAS-Bench-101 until they find the best architecture in terms of validation accuracy, whereas we limit our search to a maximal amount of $192$ queries and are able to find high-performing architectures already in this small query setting.
The comparison of AG-Net to the generator paired with an XGBoost~\cite{XGB} predictor shows that our end-to-end learnable approach is favorable even over potentially stronger predictors.
\begin{table}[t]

\caption{Results on NAS-Bench-101 for the search of the best architecture in terms of validation accuracy on CIFAR-10 to state-of-the-art methods (mean over 10 trials).}
\label{tab:NB101_Search}
\scriptsize
\begin{center}
\begin{tabular}{c||c|c||c}
\toprule
\textbf{NAS Method}  & \textbf{Val. Acc} ($\%$) & \textbf{Test Acc} ($\%$) & \textbf{Queries} \\
\midrule
\textbf{Optimum*} & $95.06$ & $94.32$& \\
\midrule
Arch2vec + RL \cite{2020Arch2vec} & - & $94.10$ & $400$\\
Arch2vec + BO \cite{2020Arch2vec}& - & $94.05$ & $400$\\
NAO \textsuperscript{\ddag}\cite{2018NAO} & $94.66$ & $93.49$ & 192 \\
BANANAS\textsuperscript{\textdagger} \cite{2021BANANAS} & $94.73$ & $94.09$ & 192 \\
Bayesian Optimization\textsuperscript{\textdagger} \cite{2015DNGO}  & $94.57$ &  $93.96$ & 192 \\
Local Search\textsuperscript{\textdagger} \cite{2020LocalSearchNAS}  & $94.57$ & $93.97$ & 192\\
Random Search\textsuperscript{\textdagger}\cite{2019RS}  & $94.31$ & $93.61$  & 192 \\
Regularized Evolution\textsuperscript{{\textdagger}}\cite{2019EvolutionaryNAS}  & $94.47$ & $93.89$ & 192 \\
WeakNAS \cite{2021WeakNAS} & - & $\textbf{94.18}$ & 200 \\
\midrule
XGB (ours) & $94.61$ & $94.13 $ & 192 \\
XGB + ranking (ours) & $94.60$ & $94.14$ & 192 \\
AG-Net (ours)  & $\textbf{94.90}$ & $\textbf{94.18}$  & 192
\\
\bottomrule
\end{tabular}
\end{center}
\end{table}

\noindent \textit{NAS-Bench-201}
This benchmark contains three different image classification tasks:
CIFAR-10, CIFAR-100 \cite{2009CIFAR} and ImageNet16-120 \cite{2017ImageNet16}.
For the experiments on NAS-Bench-201\cite{2020NB201} we retrain AG-Net in the weighted manner for $30$ epochs.
In this setting, we also compare AG-Net to two recent generative models \cite{2021GANAS,2021SGNAS}.
SGNAS \cite{2021SGNAS} trains a supernet by uniform sampling, following SETN \cite{2019SETN}. Additionally a CNN based architecture generator is trained to search architectures on the supernet.
When comparing with \cite{2020Arch2vec}, we also adopt their evaluation scheme of adding only the best-performing architecture (top-1) to the training data instead of top-16 as in our other experiments.

We report the search results for different numbers of queries for the NAS-Bench-201 dataset in \autoref{tab:NB201_Search}.
In addition, we plot the search progress in terms of queries in \autoref{fig:NB201_search} (top).
Our method provides state-of-the-art results on all datasets for a varying number of queries.
Most importantly, AG-Net shows strong performance in the few-query regime compared to \cite{2020Arch2vec} with the exception of CIFAR-100, proving its high query efficiency.

\begin{table*}[t]
\caption{Architecture Search on NAS-Bench-201. We report the mean over 10 trials for the search of the architecture with the highest validation accuracy.
}
\label{tab:NB201_Search}
\scriptsize
\begin{center}
\resizebox{\textwidth}{!}{
\begin{tabular}{c||c|c||c|c||c|c||c||c}
\toprule
\textbf{NAS Method}  & \multicolumn{2}{c||}{\textbf{CIFAR-10}} & \multicolumn{2}{c||}{\textbf{CIFAR-100}} & \multicolumn{2}{c||}{\textbf{ImageNet16-120}}  & \textbf{Queries} & \textbf{Search Method}\\
& Val. Acc  & Test Acc &Val. Acc & Test Acc &Val. Acc & Test Acc &\\
\midrule
\textbf{Optimum*} & $91.61$ & $94.37$&  $73.49$ & $73.51$& $46.77$ & $47.31$& \\
\midrule
SGNAS  \cite{2021SGNAS} & $90.18$ & $93.53$ &  $70.28$ & $70.31$ &  $44.65$ & $44.98$ & & Supernet\\
\midrule
Arch2vec + BO \cite{2020Arch2vec}  &  $91.41$  & $94.18$ &  $\textbf{73.35}$ & $\textbf{73.37}$  &  $46.34$ & $46.27$  & 100 & Bayesian Optimization \\

AG-Net (ours)& $\textbf{91.55}$ & $\textbf{94.24}$ & $73.2$ & $73.12$ & $46.31$ & $46.2$ &96 & Generative LSO\\
AG-Net (ours, topk=1) & $91.41$ & $94.16$ & $73.14$ & $73.15$ & $\textbf{46.42}$ & $\textbf{46.43}$ &100 & Generative LSO\\
\midrule

BANANAS\textsuperscript{\textdagger} \cite{2021BANANAS}  & $91.56$ & $94.3$ & $\textbf{73.49*}$ & $73.50$ & $\textbf{46.65}$& $\textbf{46.51}$ & 192 &  Bayesian Optimization \\
BO\textsuperscript{\textdagger}  \cite{2015DNGO}  & $91.54$  & $94.22$ & $73.26$ & $73.22$ &  $46.43$  & $46.40$ & 192 &  Bayesian Optimization \\
RS \textsuperscript{\textdagger} \cite{2019RS}   & $91.12$ & $93.89$ & $72.08$ & $72.07$ & $45.87$ & $45.98$ & 192  &  Random\\
XGB (ours) & $91.54$ & $94.34$ & $73.10$ & $72.93$ &  $46.48$  & $46.08$ & 192 & Generative LSO \\
XGB + Ranking (ours) & $91.48$ & $94.25$ & $73.20$ & $73.24$ & $46.40$ & $46.16$ & 192 &Generative LSO \\
AG-Net (ours) & $\textbf{91.60}$ & $\textbf{94.37*}$ & $\textbf{73.49*}$ & $\textbf{73.51*}$ & $46.64$ & $46.43$ &192 &Generative LSO \\
\midrule
GANAS \cite{2021GANAS}& - & $94.34$   & - & $73.28$ &  - & $\textbf{46.80}$ & 444  & Generative Reinforcement Learning\\
AG-Net (ours) & $\textbf{91.61*}$ & $\textbf{94.37*}$   &  $\textbf{73.49*}$ & $\textbf{73.51*}$ &  $\textbf{46.73}$ & $46.42$ & 400 &Generative LSO \\
\bottomrule
\end{tabular}}
\end{center}
\end{table*}
\setlength{\tabcolsep}{4pt}


\subsection{Experiments on Surrogate Benchmarks}\label{sec:surr_benchmark_experiments}
We furthermore apply our search method on larger search spaces as DARTS \cite{2018DARTS} and NAS-Bench-NLP \cite{2020NBNLP} without ground truth evaluations for the whole search space, making use of surrogate benchmarks as NAS-Bench-301 \cite{2020NB301}, NAS-Bench-X11 \cite{2021NBX11} and NAS-Bench-Suite \cite{NBSuite}.

\begin{figure*}[t]
    \centering
    \includegraphics[width=0.3\textwidth]{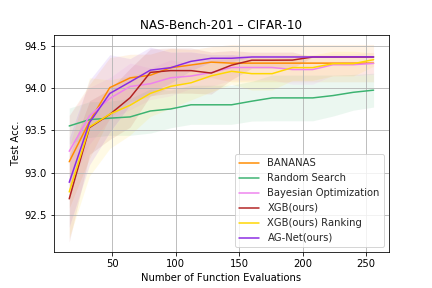}
    \includegraphics[width=0.3\textwidth]{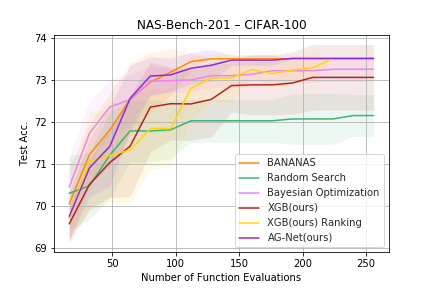}
    \includegraphics[width=0.3\textwidth]{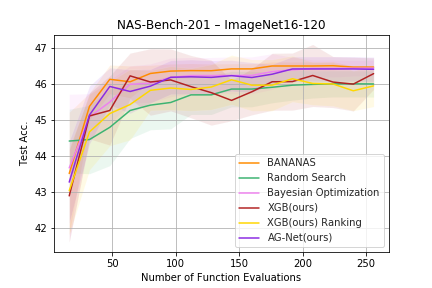} \\
    \includegraphics[width=0.3\textwidth]{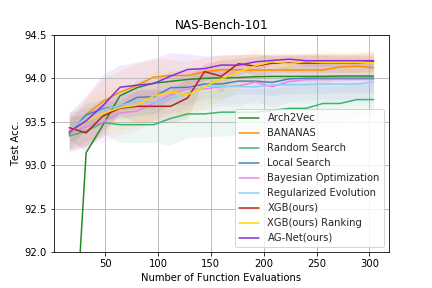}
    \includegraphics[width=0.3\textwidth]{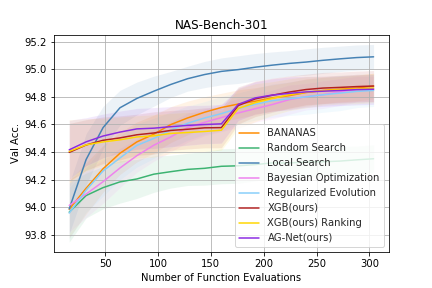}
    \includegraphics[width=0.3\textwidth]{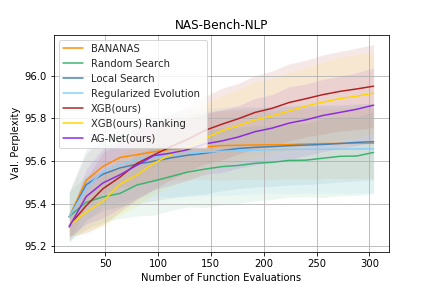}

    \caption{
    Architecture search evaluations on NAS-Bench-201, NAS-Bench-101, NAS-Bench-301 and NAS-Bench-NLP for different search methods.
    \label{fig:NB201_search} \label{fig:NB101_search} \label{fig:NB201_search} \label{fig:NB301_search}}
\end{figure*}

\noindent \textit{NAS-Bench-301}
Here, we report experiments on the cell-based DARTS \cite{2018DARTS} search space using the surrogate benchmark NAS-Bench-301 \cite{2020NB301} for the CIFAR-10 \cite{2009CIFAR} image classification task.
The exact search procedure using the cells individually is described in the supp. mat. \autoref{sec:supp_darts}.
The results are described in \autoref{tab:NB301_Search} (left) and visualized in \autoref{fig:NB301_search} (bottom middle). Our method is comparable to other state-of-the-art methods in this search space. 

\begin{table}[t]
\caption{Results on: (\textbf{left}) NAS-Bench-301 (mean validation accuracy over 50 trials). (\textbf{right}) NAS-Bench-NLP (mean validation perplexity over 100 trials).}
\label{tab:NBNLP_Search}\label{tab:NB301_Search}
\scriptsize
\begin{center}
\begin{tabular}{c||c|c||c|c}
\toprule
\textbf{NAS Method}  & \multicolumn{2}{c||}{\textbf{NAS-Bench-301}} & \multicolumn{2}{c}{\textbf{NAS-Bench-NLP}} \\
& Val. Acc ($\%$) & Queries & Val. Perplexity ($\%$) & Queries \\
\midrule
BANANAS\textsuperscript{\textdagger} \cite{2021BANANAS} & $94.77$  & 192 & $95.68$  & 304 \\
Bayesian Optimization\textsuperscript{\textdagger}  \cite{2015DNGO}  & $94.71$ & 192  & - & -  \\
Local Search\textsuperscript{\textdagger} \cite{2020LocalSearchNAS} & $\textbf{95.02}$ & 192  & $95.69$ & 304 \\
Random Search\textsuperscript{\textdagger}\cite{2019RS}  & $94.31$ & 192  & $95.64$  & 304 \\
Regularized Evolution\textsuperscript{{\textdagger}}\cite{2019EvolutionaryNAS}  & $94.75$ & 192  & $95.66$ & 304 \\
\midrule
XGB (ours)  & $94.79$  &  192 & $\textbf{95.95}$ & 304
\\
XGB + Ranking (ours) & $94.76$ &  192 & $95.92$  & 304
\\
\midrule
AG-Net (ours)  & $94.79$ &  192 &  $95.86$ & 304
\\
\bottomrule
\end{tabular}
\end{center}
\end{table}

\noindent \textit{NAS-Bench-NLP}
Next, we evaluate AG-Net on NAS-Bench-NLP \cite{2020NBNLP} for the language modeling task on Penn TreeBank \cite{2010PTB}. We retrain AG-Net coupled with the surrogate model for 30 epochs to predict the validation perplexity. Note, since the search space considered in NAS-Bench-NLP is too large for a full tabular benchmark evaluation, we make use of the surrogate benchmark NAS-Bench-X11 \cite{2021NBX11} and NAS-Bench-Suite \cite{NBSuite} instead of tabular entries.

For fair comparison we compare our methods to the same state-of-the-art methods as in previous experiments. 
The results are reported in \autoref{tab:NBNLP_Search} (right) and visualized in \autoref{fig:NB301_search} (bottom right). Our AG-Net improves over all state-of-the-art methods by a substantial margin and using XGB as a predictor even improves the search further.

\noindent \textbf{ImageNet Experiments} 
\begin{table}[t]
\caption{ImageNet \textbf{error} of neural architecture search on DARTS.}
\label{tab:DARTS}
\scriptsize
\begin{center}
\begin{tabular}{c||c|c||c || c}
\toprule
\textbf{NAS Method}  & \textbf{Top-1}\textdownarrow & \textbf{Top-5}\textdownarrow & \textbf{\# Queries} & \makecell{\textbf{Search} \\ GPU days} \\
\midrule
\multicolumn{1}{c}{{Mixed Methods}} & \multicolumn{4}{c}{} \\
\midrule
NASNET-A (CIFAR-10) \cite{2018LearningNAS} & $26.0$ & $8.4$ & 20000 & 2000 \\
PNAS (CIFAR-10) \cite{2018PNAS} & $25.8$ & $8.1$ & 1160 & 225\\
NAO (CIFAR-10) \cite{2018NAO} & $24.5$ & $7.8$ & 1000 & 200 \\
\midrule
\multicolumn{1}{c}{{Differentiable Methods}} & \multicolumn{4}{c}{} \\
\midrule
DARTS (CIFAR-10) \cite{2018DARTS} & $26.7$ & $8.7$ & - & 4.0 \\
SNAS (CIFAR-10)\cite{2019SNAS} & $27.3$ & $9.2$ & - & 1.5\\
PDARTS (CIFAR-10) \cite{19PDARTS} & $\textbf{24.4}$ & $\textbf{7.4}$ & - & 0.3 \\
PC-DARTS (CIFAR-10) \cite{20PCDARTS} & $25.1$ & $7.8$ & - & {\textbf{0.1}} \\
PC-DARTS (ImageNet) \cite{20PCDARTS} & $\mathbf{24.2}$ & $\mathbf{7.3}$ & - & 3.8 \\
 \midrule
\multicolumn{1}{c}{{Predictor Based Methods}} & \multicolumn{4}{c}{} \\
\midrule
WeakNAS (ImageNet) \cite{2021WeakNAS} &\textbf{ 23.5} &\textbf{ 6.8} & 800 & 2.5 \\
XGB (NB-301)(CIFAR-10) (ours) & 24.1 & 7.4 & 304 & \textbf{0.02} \\
XGB + Ranking (NB-301)(CIFAR-10) (ours) & 24.1 & 7.2 & 304 & \textbf{0.02} \\
AG-Net (NB-301)(CIFAR-10) (ours) & 24.3 & 7.3 & 304 & 0.21 \\
\midrule
\multicolumn{1}{c}{{Training-Free Methods}} & \multicolumn{4}{c}{} \\
\midrule
TE-NAS (CIFAR-10)\cite{2021Im4GPU} & $26.2$ & $8.3$ & - & 0.05 \\
TE-NAS (ImageNet)\cite{2021Im4GPU} & $24.5$ & $7.5$ & - & 0.17 \\
AG-Net (CIFAR-10) (ours)  & $\textbf{23.5}$ & $7.1$  & 208 & \textbf{0.02} \\
AG-Net (ImageNet) (ours)  & $\textbf{23.5}$ & $6.9$  & 208 & \textbf{0.09}
\\
\bottomrule
\end{tabular}
\end{center}
\end{table}
The previous experiment on NAS-Bench-301 \cite{2020NB301} shows the ability of our generator to generate valid architectures and to perform well in the DARTS \cite{2018DARTS} search space. This 
allows for searching a well-performing architecture on ImageNet \cite{2009ImageNet}. Yet evaluating up to 300 different found architectures on ImageNet is extremely expensive. Our first approach is to retrain the best found architectures on the CIFAR-10 \cite{2009CIFAR} image classification task from the previous experiment on NAS-Bench-301 (AG-Net and the XGBoost adaptions) on ImageNet \cite{2009ImageNet}. Our second approach is based on a training-free neural architecture search approach. The recently proposed TE-NAS \cite{2021Im4GPU} provides a training-free neural architecture search approach, by ranking architectures by analysing the neural tangent kernel (NTK) and the number of linear regions (NLR) of each architecture. These two measurements are training free and do not need any labels. The intuition between those two measurements is their implication towards trainability and expressivity of a neural architecture and also their correlation with the neural architecture's accuracy; NTK is negatively correlated and NLR positively correlated with the architecture's test accuracy. We adapt this idea for our search on ImageNet and search architectures in terms of their NTK value and their number of linear regions instead of their validation accuracy. We describe the detailed search process in the supp. mat. \autoref{sec:supp_darts}.

\autoref{tab:DARTS} shows the results.
Note that our latter described search method on ImageNet is \textbf{training-free} (as TE-NAS \cite{2021Im4GPU}) and the amount of queries displays the amount of data we evaluated for the zero cost measurements. Other query information include the amount of (partly) trained architectures. 
Furthermore, the displayed differentiable methods are based on training supernets which can lead to expensive training times.  The best found architectures on NAS-Bench-301 \cite{2020NB301} (CIFAR-10) result in comparable error rates on ImageNet to former approaches.
As a result, our search method approach is highly efficient and outperforms previous methods in terms of needed GPU days. 
The result in terms of top-1 and top-5 error rates are even improving over the one from previous approaches when using the training free approach.


{\begingroup 
\setlength\tabcolsep{4.5pt} 
\begin{table*}[t]
            \caption{
Results for searches with at most $200$ queries on HW-NAS-Bench \cite{2021HWNNB} with varying devices and latency (Lat.) constraints in two multi-objective settings:
\emph{Joint=0} optimizes accuracy under latency constraint, while \emph{Joint=1} optimizes for accuracy and latency jointly.
We report the best found architecture out of $10$ runs with their corresponding latency, as well as the mean of these runs.
We compare to random search as a strong baseline \cite{2019RS}.
Feasibility (Feas.) is the proportion of evaluated architectures during the search that satisfy the latency constraint (larger is better).
The optimal architecture ($\textbf{*}$) is the architecture with the highest accuracy satisfying the latency constraint.
            }
        \label{tab:hwbench-results}
    \begin{center}\begin{scriptsize}
    \resizebox{\textwidth}{!}{
        \begin{tabular}{cc||cc|cc|cc||cc|cc|cc||cc}
        \toprule
        \multicolumn{2}{c||}{\textbf{Settings}} & \multicolumn{6}{c||}{\textbf{Best out of 10 runs}} & \multicolumn{6}{c||}{\textbf{Mean}} \\
        \multicolumn{2}{c||}{Constraint} &
        \multicolumn{2}{c|}{\textbf{Joint=0}} & \multicolumn{2}{c|}{\textbf{Joint=1}} & \multicolumn{2}{c||}{\textbf{Random}} & \multicolumn{2}{c|}{\textbf{Joint=0}} & \multicolumn{2}{c|}{\textbf{Joint=1}} & \multicolumn{2}{c||}{\textbf{Random}} & \multicolumn{2}{c}{\textbf{Optimum}*} \\
        Device & Lat.\textdownarrow & Acc.\textuparrow & Lat.\textdownarrow & Acc.\textuparrow & Lat.\textdownarrow & Acc.\textuparrow & Lat.\textdownarrow & Acc.\textuparrow & Feas.\textuparrow & Acc.\textuparrow & Feas.\textuparrow & Acc.\textuparrow & Feas.\textuparrow & Acc.\textuparrow & Lat.\textdownarrow \\
        \midrule
Edge GPU & $2$ & $\textbf{0.406*}$ & $1.90$ & $\textbf{0.406*}$ & $1.90$ & $0.397$ & $1.78$ & $\textbf{0.397}$ & $0.29$ & $\textit{0.391}$ & $0.31$ & $0.372$ & $0.05$ & $0.406$ & $1.90$ \\
Edge GPU & $4$ & $\textbf{0.448*}$ & $3.49$ & $\textbf{0.448*}$ & $3.49$ & $0.437$ & $3.35$ & $\textit{0.428}$ & $0.29$ & $\textbf{0.433}$ & $0.43$ & $0.417$ & $0.22$ & $0.448$ & $3.49$ \\
Edge GPU & $6$ & $\textit{0.458}$ & $5.29$ & $\textbf{0.464*}$ & $5.96$ & $\textit{0.458}$ & $5.29$ & $\textbf{0.453}$ & $0.64$ & $\textit{0.450}$ & $0.79$ & $0.449$ & $0.72$ & $0.464$ & $5.96$ \\
Edge GPU & $8$ & $\textit{0.465}$ & $6.81$ & $\textbf{0.468*}$ & $6.81$ & $0.464$ & $7.44$ & $\textbf{0.463}$ & $0.98$ & $\textit{0.462}$ & $0.99$ & $0.457$ & $1.00$ & $0.468$ & $6.81$ \\
        \midrule
Raspi 4 & $2$ & $\textbf{0.355*}$ & $1.58$ & $\textbf{0.355*}$ & $1.58$ & $0.348$ & $1.60$ & $\textit{0.346}$ & $0.28$ & $\textbf{0.347}$ & $0.30$ & $0.339$ & $0.08$ & $0.355$ & $1.58$ \\
Raspi 4 & $4$ & $\textit{0.431}$ & $3.83$ & $\textbf{0.436*}$ & $3.79$ & $0.427$ & $3.85$ & $\textit{0.420}$ & $0.47$ & $\textbf{0.428}$ & $0.50$ & $0.419$ & $0.37$ & $0.436$ & $3.79$ \\
Raspi 4 & $6$ & $\textit{0.449}$ & $5.95$ & $\textbf{0.452*}$ & $5.29$ & $0.445$ & $5.95$ & $\textit{0.440}$ & $0.56$ & $\textbf{0.441}$ & $0.57$ & $0.432$ & $0.55$ & $0.452$ & $5.29$ \\
Raspi 4 & $8$ & $\textit{0.456}$ & $6.33$ & $0.455$ & $7.96$ & $\textbf{0.457}$ & $7.97$ & $\textbf{0.451}$ & $0.69$ & $\textit{0.449}$ & $0.79$ & $0.447$ & $0.76$ & $0.465$ & $7.43$ \\
Raspi 4 & $10$ & $\textbf{0.466}$ & $8.66$ & $\textit{0.465}$ & $8.62$ & $0.464$ & $8.72$ & $\textbf{0.464}$ & $0.77$ & $\textit{0.454}$ & $0.94$ & $\textit{0.454}$ & $0.90$ & $0.468$ & $8.83$ \\
Raspi 4 & $12$ & $\textbf{0.468*}$ & $8.83$ & $0.463$ & $9.05$ & $\textit{0.464}$ & $8.72$ & $\textbf{0.465}$ & $0.91$ & $\textit{0.457}$ & $0.98$ & $0.456$ & $0.96$ & $0.468$ & $8.83$ \\
        \midrule
Edge TPU & $1$ & $\textbf{0.468*}$ & $0.96$ & $\textit{0.466}$ & $0.97$ & $0.464$ & $1.00$ & $\textbf{0.464}$ & $0.74$ & $\textit{0.457}$ & $0.82$ & $0.454$ & $0.79$ & $0.468$ & $0.96$ \\
        \midrule
Pixel 3 & $2$ & $\textbf{0.413*}$ & $1.30$ & $\textbf{0.413*}$ & $1.30$ & $0.400$ & $1.50$ & $\textbf{0.409}$ & $0.48$ & $\textit{0.405}$ & $0.59$ & $0.388$ & $0.30$ & $0.413$ & $1.30$ \\
Pixel 3 & $4$ & $\textbf{0.460*}$ & $3.55$ & $0.446$ & $3.01$ & $\textit{0.447}$ & $3.23$ & $\textbf{0.453}$ & $0.69$ & $\textit{0.441}$ & $0.77$ & $0.438$ & $0.64$ & $0.460$ & $3.55$ \\
Pixel 3 & $6$ & $\textit{0.464}$ & $5.92$ & $\textbf{0.465*}$ & $5.95$ & $0.458$ & $4.68$ & $\textbf{0.457}$ & $0.77$ & $\textit{0.452}$ & $0.94$ & $0.451$ & $0.88$ & $0.465$ & $5.57$ \\
Pixel 3 & $8$ & $\textbf{0.468*}$ & $6.65$ & $\textit{0.465}$ & $7.88$ & $0.461$ & $7.13$ & $\textbf{0.464}$ & $0.87$ & $\textit{0.457}$ & $0.99$ & $0.454$ & $0.97$ & $0.468$ & $6.65$ \\
Pixel 3 & $10$ & $\textbf{0.466}$ & $6.70$ & $0.461$ & $8.48$ & $\textit{0.464}$ & $8.01$ & $\textbf{0.464}$ & $0.96$ & $0.455$ & $1.00$ & $\textit{0.456}$ & $0.99$ & $0.468$ & $6.65$ \\
        \midrule
Eyeriss & $1$ & $\textbf{0.452*}$ & $0.98$ & $\textit{0.449}$ & $0.98$ & $0.447$ & $0.98$ & $\textbf{0.445}$ & $0.49$ & $\textit{0.436}$ & $0.53$ & $0.433$ & $0.23$ & $0.452$ & $0.98$ \\
Eyeriss & $2$ & $\textbf{0.465}$ & $1.65$ & $\textbf{0.465}$ & $1.65$ & $0.464$ & $1.65$ & $\textbf{0.463}$ & $0.87$ & $\textit{0.457}$ & $0.99$ & $\textit{0.457}$ & $0.95$ & $0.468$ & $1.65$ \\
        \midrule
FPGA & $1$ & $\textbf{0.440}$ & $1.00$ & $\textbf{0.440}$ & $0.97$ & $0.438$ & $0.97$ & $\textbf{0.433}$ & $0.65$ & $\textbf{0.433}$ & $0.80$ & $0.429$ & $0.58$ & $0.444$ & $1.00$ \\
FPGA & $2$ & $\textbf{0.465*}$ & $1.60$ & $0.460$ & $1.60$ & $\textit{0.463}$ & $1.97$ & $\textbf{0.462}$ & $0.82$ & $0.451$ & $0.99$ & $\textit{0.453}$ & $0.97$ & $0.465$ & $1.60$ \\
        \bottomrule
        \end{tabular}}
    \end{scriptsize}\end{center}
\end{table*}
\endgroup}

\begin{figure}[t]
    \centering
    \includegraphics[width=0.3\textwidth]{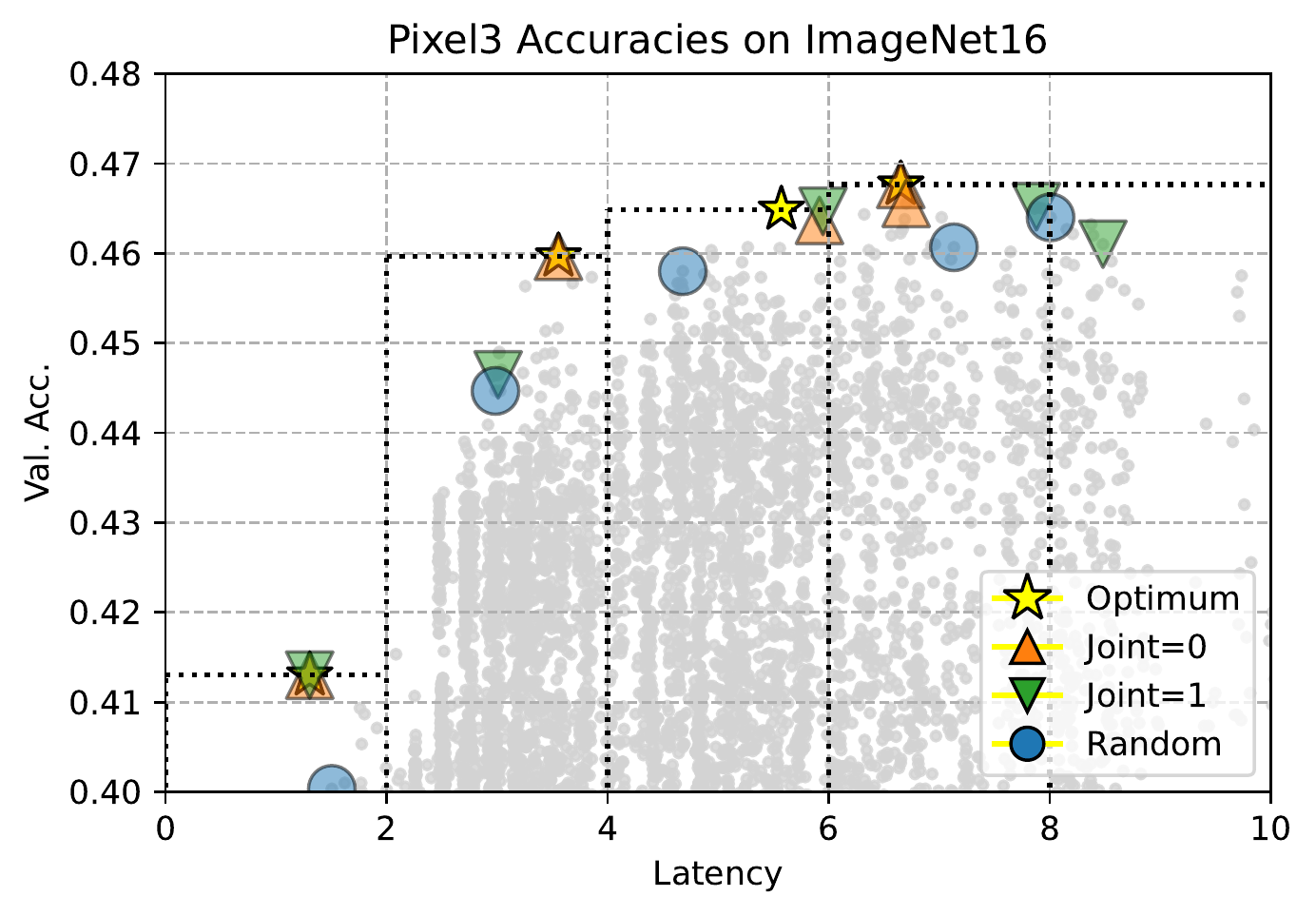}
     \includegraphics[width=0.3\textwidth]{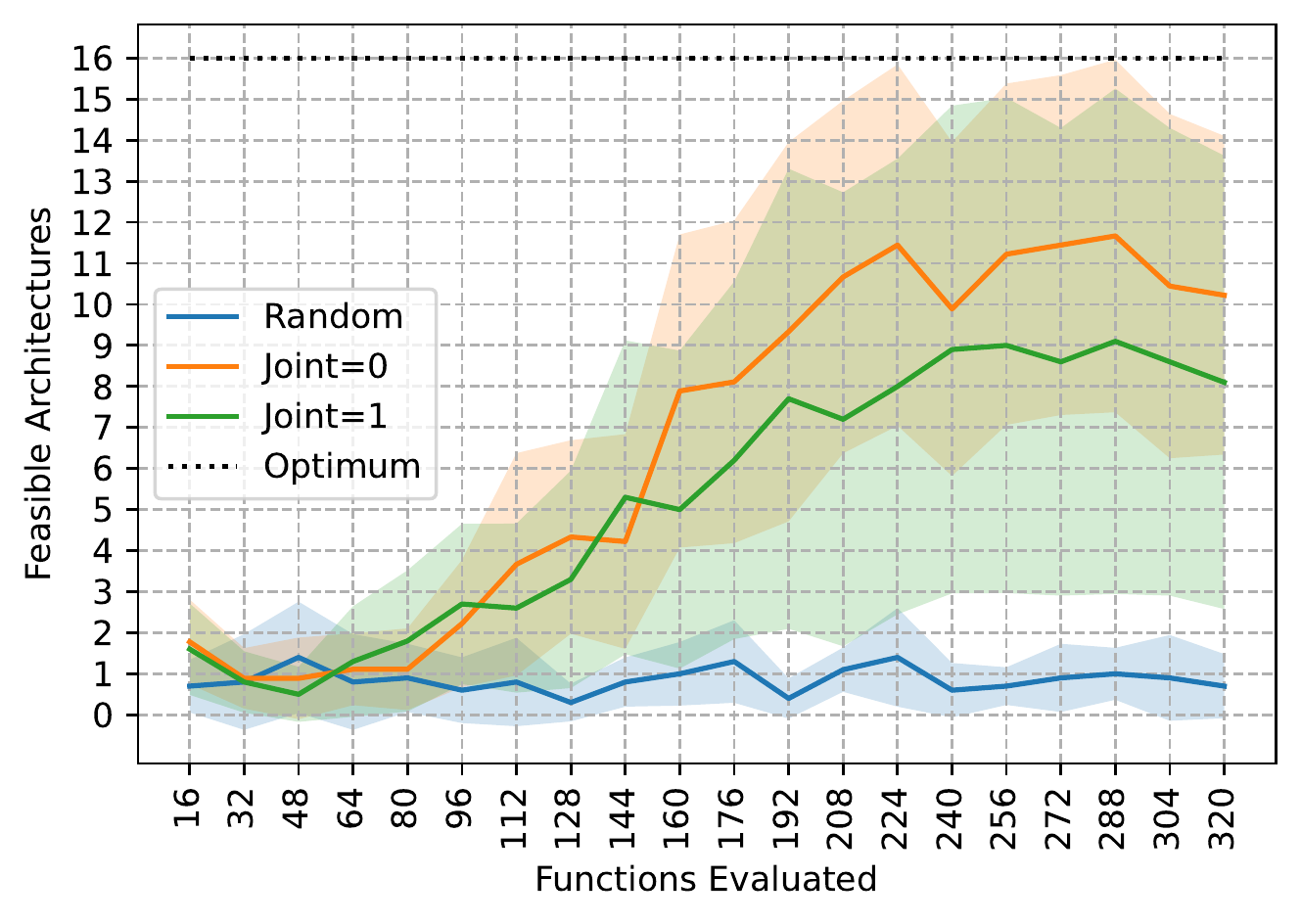}

    \caption{ (\textbf{left})
    Exemplary searches on HW-NAS-Bench for image classification on ImageNet16 with $192$ queries on Pixel 3 and latency conditions $L\in\{2,4,6,8,10\}$ (y-axis zoomed for visibility). (\textbf{right}) Amount of architectures generated and selected in each search iteration (at most $16$) that satisfy the latency constraint.  In this example we searched on Edge GPU with $L=2$.
    \label{fig:HW_search} \label{fig:feasibility}}
\end{figure}

\subsection{Experiments on Hardware-Aware Benchmark}
\label{sec:hwbench-experiments}
Next, we apply AG-Net to the Hardware-Aware NAS-Benchmark \cite{2021HWNNB}.
We demonstrate in two settings that AG-Net can be used for multi-objective learning.
The first setting (\emph{Joint=1}) is formulated as constrained joint optimization:

\begin{align}\label{eq:joint=1_optimization}
     & \underset{G \sim p_D}{\max}~ f(G) \wedge \underset{G \sim p_D,  }{\min} ~ g_h(G) 
     &
     & \textrm{s.t.}~  g_h(G) \leq L,\exists~ h \in H, 
\end{align}
where $f(\cdot)$ evaluates architecture $G$ for accuracy and $g_h(\cdot)$ evaluates for latency given a hardware $h \in H$ and a user-defined latency constraint $L$.
The second setting (\emph{Joint=0}) is formulated as constraint objective:

\begin{align}\label{eq:joint=0_optimization}
    & \underset{G \sim p_D}{\max} ~ f(G)  &
    & \textrm{s.t.}~  g_h(G) \leq L,\exists~ h \in H, 
\end{align}
where we drop the optimization on latency and only optimize accuracy given the latency constraint.
The loss function to train our generator in these settings is updated from \autoref{eq:search_loss} to:
\begin{align}\label{eq:hw_cond_loss}
\begin{split}
\mathcal{L}(\tilde{G},G) = & (1-\alpha) \mathcal{L}_G(\tilde{G},G) + \alpha \big[\lambda \mathcal{L}_{C_1}(\tilde{G},G)+ (1- \lambda)\mathcal{L}_{C_2}(\tilde{G},G)\big], 
        \end{split}
\end{align}
where $\alpha$ is a hyperparameter trading off generation and prediction loss, and $\lambda$ is a hyperparameter trading off both prediction targets $C_1$ (accuracy) and $C_2$ (latency).

To perform LSO in the joint objective setting from \autoref{eq:joint=1_optimization}, we rank the training data $D$ for both accuracy and latency jointly by summing both individual rankings.
To fulfill the optimization constraint, we further penalize the ranks via a multiplicative penalty if the latency does not fulfill the constraint.
This overall ranking is then used for the weight calculation in \autoref{eq:weights}.
The LSO for the constraint objective setting from \autoref{eq:joint=0_optimization} only ranks architectures by accuracy and penalizes architectures with infeasible latency property.
We choose random search as a baseline in this setting as it is generally regarded as a strong baseline in NAS \cite{2019RS}.
\autoref{fig:HW_search} depicts searches with our model in both optimization settings on Pixel 3 with different latency conditions.
More results on different hardware and latency constraints are shown in \autoref{tab:hwbench-results}.
We observe that either optimization setting outperforms the random search baseline in almost all tasks.
Additionally, our method is able to find the optimal architecture for a task regularly (in $15$ out of $20$ tasks), which random search was not able to provide.
When considering mean accuracy and feasibility of the best architectures of all runs, we see that \emph{Joint=1} is able to improve the ratio of feasible architectures found during the search substantially.
This is to be expected given that the latent space is explicitly optimized for latency in this setting.
Consequently, \emph{Joint=1} is able to find better-performing architectures compared to \emph{Joint=0} if the constraint restricts the space of feasible architectures strongly (see results on Raspi 4).
The feasibility ratio of random search is an indicator on how restricted the space is.
In most cases, the latency penalization seems to be sufficient to find enough well-performing and feasible architectures, as can be seen by the feasibility of \emph{Joint=0} which is greatly improved compared to random search.
We show the development of feasibility over time from \autoref{tab:hwbench-results} in \autoref{fig:feasibility}.


\begin{table*}[t]
\caption{Ablation: Search results on NAS-Bench-101 and NAS-Bench-201 using AG-Net (mean over 10 trials with a maximal query amount of $192$).}
\label{tab:NB101201_Search_ablation_LSO}
\scriptsize
\begin{center}
\resizebox{\textwidth}{!}{
\begin{tabular}{c||c|c||c|c|c|c|c|c}
\toprule
 & \multicolumn{2}{c||}{\textbf{NAS-Bench-101}} & \multicolumn{6}{c}{\textbf{NAS-Bench-201}} \\
&  \multicolumn{2}{c||}{CIFAR-10}& \multicolumn{2}{c|}{CIFAR-10} & \multicolumn{2}{c|}{CIFAR-100} & \multicolumn{2}{c}{ImageNet16-120}    \\
& Val. Acc  & Test Acc & Val. Acc  & Test Acc &Val. Acc & Test Acc &Val. Acc & Test Acc \\
\midrule
\textbf{Optimum}* &$95.06$ & $94.32$ & $91.61$ & $94.37$&  $73.49$ & $73.51$& $46.77$ & $47.31$ \\
\midrule

AG-Net (ours) w/o LSO & $94.38$ & $93.78$&  $91.15$ & $93.84$ & $71.72$  &  $71.83$ &  $45.33$ & $45.04$  \\
AG-Net (ours) w/o backprop & $94.71$ & $94.12$& $\textbf{91.60}$  & $94.30$ & $73.38$  & $73.22$  & $46.62$  & $46.13$\\
AG-Net (ours) & $\textbf{94.90}$ & $\textbf{94.18}$& $\textbf{91.60}$ & $ \textbf{94.37*}$ & $\textbf{73.49*}$ & $\textbf{73.51*}$ & $\textbf{46.64}$ & $\textbf{46.43}$  \\
\bottomrule
\end{tabular}}
\end{center}
\end{table*}

\subsection{Ablation Studies}\label{sec:ablation_studies}

In this section we analyse the impact of the LSO technique and the backpropagation ability to the search efficiency. Therefore, we compare our AG-Net with the latter named adaptions on the tabular benchmarks NAS-Bench-101 \cite{2019NB101} and NAS-Bench-201 \cite{2020NB201}. The results of our ablation study are reported in \autoref{tab:NB101201_Search_ablation_LSO}.
As we can see, the lack of weighted retraining decreases the search substantially. 
In addition the results without backpropagation support that the coupling of the predictor's target and the generation process enables a more efficient architecture search over different search spaces. Thus, the combination of LSO and a fully differentiable approach improves the effectiveness of the search.


\section{Conclusion}
We propose a simple architecture generative network (AG-Net), which allows us to directly generate architectures without any additional encoder or discriminator. AG-Net is fully differentiable, allowing to couple it with surrogate models for different target predictions.
In contrast to former works, it enables to backpropagate the target information from the surrogate predictor into the generator.
By iteratively optimizing the latent space of the generator, our model learns to focus on promising regions of the architecture space, so that it can generate high-scoring architectures directly in a query and sample-efficient manner.
Extensive experiments on common NAS benchmarks demonstrate that our model outperforms state-of-the-art methods at almost any time during architecture search and achieves state-of-the-art performance on ImageNet.
It also allows for multi-objective optimization on the Hardware-Aware NAS-Benchmark.

\subsubsection{Acknowledgments.}
JL and MK acknowledge the German Federal Ministry of Education and Research Foundation via the project DeToL.

%
\bibliographystyle{splncs04}
\bibliography{egbib, eccv_2022}

\onecolumn
\newpage
\appendix
\section*{Appendices}

Section \ref{sup:sp-representations} provides an overview about the graph representations for each search space, we consider in the main paper.
In \autoref{supp:sec_ablation} we show additional ablation studies.
In \autoref{sup:implementation-details}, we provide more details about the experimental settings. 
In \autoref{sec:supp_darts} we 
provide additional details for our search method on the DARTS search space.
In \autoref{sec:generator} we describe details about the generator network, and in \autoref{sec:HP} we list all hyperparameter settings of our experiments.
Lastly, we include a visual intuition of the latent space optimization technique in \autoref{sec:lso_intuition}.

\section{Search Space Representations}\label{sup:sp-representations}
In this section we give more details about the search spaces we consider in the main paper. 
\subsection{NAS-Bench-101}

\begin{figure*}
	\centering
	\includegraphics[height=5cm]{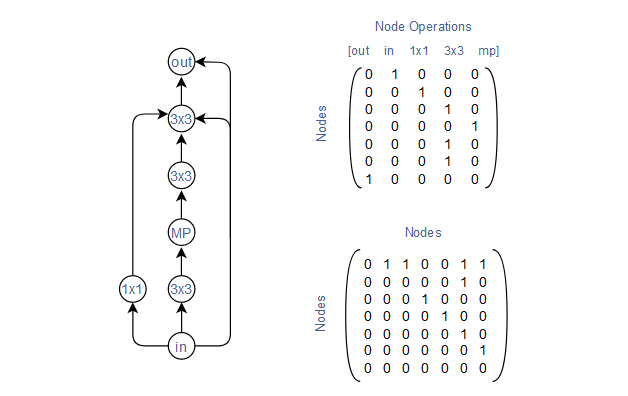}
	
	\caption{Exemplary cell representation from the NAS-Bench-101 search space. (\textbf{left}) DAG representation of a graph with 7 nodes. (\textbf{right}) The top part shows the node attribute matrix to the DAG and the bottom part shows its adjacency matrix. \label{fig:NB101_representation}}
	
\end{figure*}
NAS-Bench-101 is the first tabular benchmark designed for benchmarking NAS methods.
This search space is a cell-based search space and contains $423,624$ unique neural networks.
Each architecture is trained $3$ times on CIFAR-10 \cite{2009CIFAR} for image classification.
The cell topology is limited to the number of nodes $\vert V \vert \leq 7$ (including input and output nodes) and edges $\vert E \vert \leq 9$.
The nodes represent the architecture layers and intermediate nodes can take any operation from the operation set $\mathcal{O} = \{1 \times 1~\textrm{conv.}, 3 \times 3~\textrm{conv.}, 3 \times 3~\textrm{max pooling}\}$.
For visualization purposes, we present in \autoref{fig:NB101_representation} exemplary a DAG from the NAS-Bench-101 search space, with its corresponding node attribute matrix and its adjacency matrix. Note, a concatenation of the flatted node attribute matrix and the flatted upper triangular adjacency matrix is the representation our generator model is trained to learn; this holds for all search spaces.

\subsection{NAS-Bench-201}
\begin{figure*}
	\centering
	\includegraphics[height=7cm]{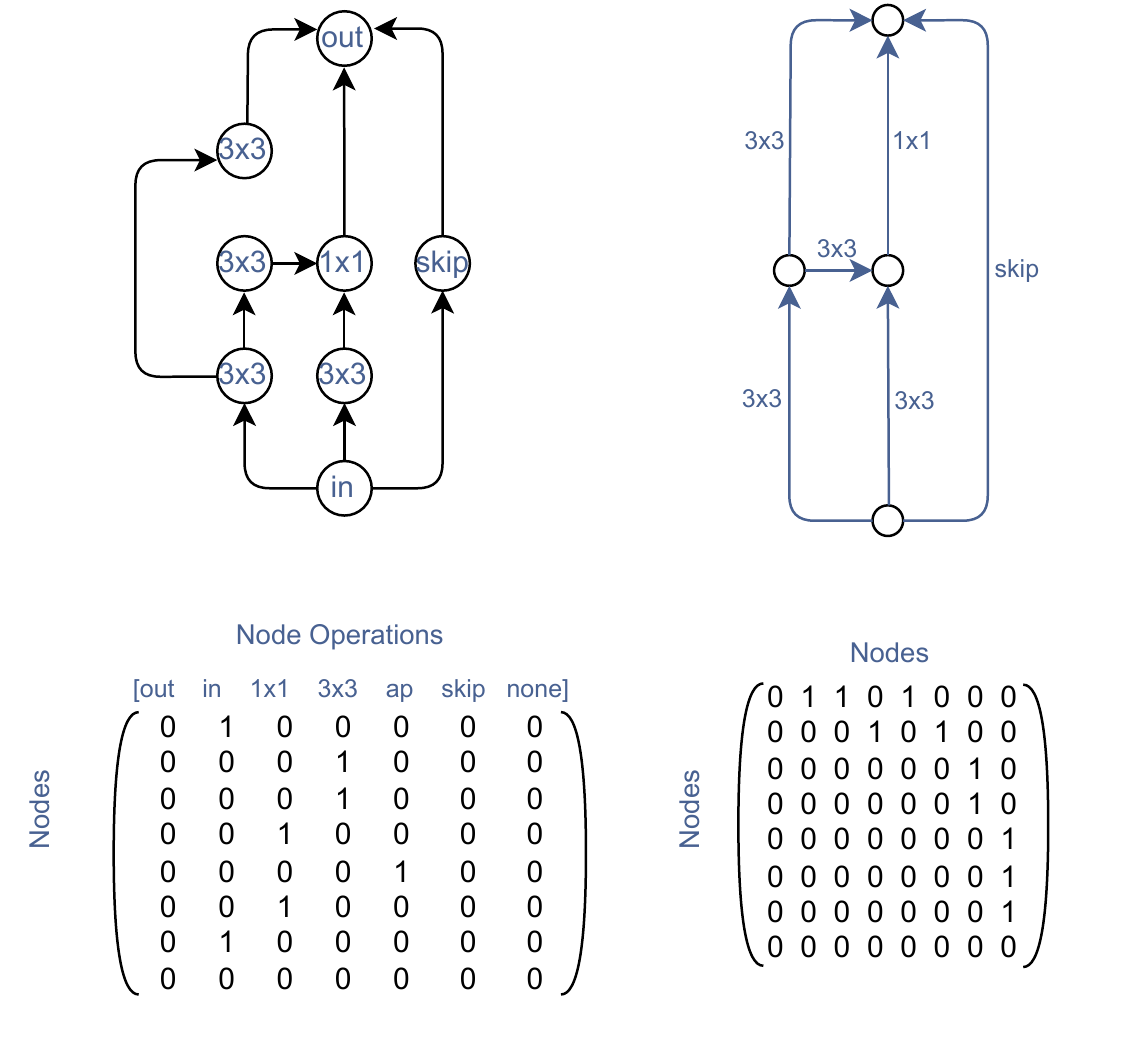}
	\caption{Exemplary cell representation from the NAS-Bench-201 search space. (\textbf{top}) The left part visualizes the DAG representation with node attributes instead of edge attributes. The right part shows the true DAG representation in the NAS-Bench-201 search space. (\textbf{bottom}) The left part shows the node attribute matrix to the DAG and the right part shows its adjacency matrix. \label{fig:NB201_representation}}
\end{figure*}

NAS-Bench-201 \cite{2020NB201} is another 
cell-structured search space, which consists of $15,625$ architectures.
Each architecture is trained for $200$ training epochs
on CIFAR-10 \cite{2009CIFAR}, CIFAR-100 \cite{2009CIFAR}, and ImageNet16-120 \cite{2017ImageNet16}.
This benchmark provides validation and test accuracy information for each of the three datasets.
The cell structure is different compared to NAS-Bench-101: Each cell has $\vert V \vert = 4$ nodes and $\vert E \vert = 6$ edges, where the former represent
feature maps and the latter denote operations chosen from the set $\mathcal{O} = \{1 \times 1~\textrm{conv.}, 3 \times 3~\textrm{conv.}, 3 \times 3~\textrm{avg pooling}, \textrm{skip}, \textrm{zero}\}$.

\autoref{fig:NB201_representation} visualizes a DAG in the true variant in the NAS-Bench-201 search space with edge attributes, as well as our adapted representation, where the edge attributes are changed to node attributes. This is similar to the representation in \cite{2020Arch2vec}.
We show experiments on NAS-Bench-101 and NAS-Bench-201 in \autoref{sec:tabluar_benchmark_experiments}.

\subsection{DARTS Search Space}
\begin{figure*}
	\centering
	\includegraphics[height=7cm]{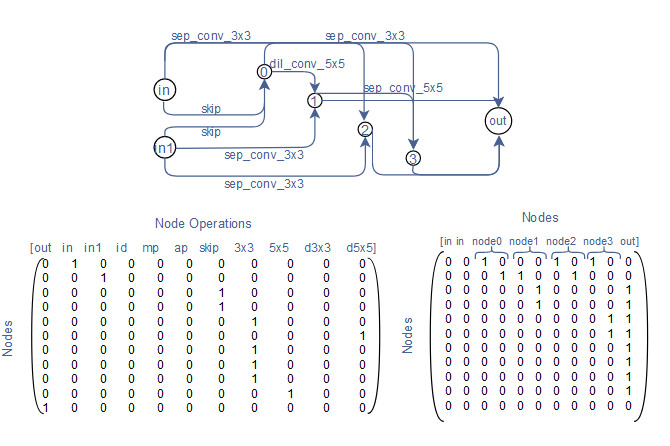}
	\caption{Exemplary cell representation from the DARTS search space. (\textbf{top}) Visualization of the DAG representation in the DARTS search space. (\textbf{bottom}) The left part shows the node attribute matrix to the DAG and the right part shows its adjacency matrix. \label{fig:NB301_representation}}
\end{figure*}
NAS-Bench-301 \cite{2020NB301} is the first surrogate benchmark, which evaluates several surrogate models on in total $60,000$ sampled architectures from the DARTS \cite{2018DARTS} search space on the CIFAR-10 \cite{2009CIFAR}
image classification task. 
The DARTS search space consists of $10^{18}$ neural networks, where each network consists of two cells; a normal cell and a reduction cell.  Each cell is limited by the number of nodes $\vert N \vert = 7$ and the number of edges $\vert E \vert = 12$, where $4$ of these edges connect the intermediate nodes (excluding the input nodes) to the output node. Each edge denotes an operation from the set  $\mathcal{O} = \{3 \times 3~\textrm{sep. conv.}, 5 \times 5~\textrm{sep. conv.}, 3 \times 3~\textrm{dil. conv.}, 5 \times 5~\textrm{dil. conv.},  3 \times 3~\textrm{avg pooling}, 3 \times 3~\textrm{max pooling}, \textrm{identity}, \textrm{zero}\}$. Each intermediate edge is connected to two predecessor nodes. Each cell also contains two input nodes, which are the output nodes from the previous two cells. The overall network is created by stacking the normal and reduction cell.

In order to train our generative model to generate valid cells, we additionally randomly sample $500$k architectures from the DARTS search space. 
We train our generative model to learn to generate valid cells independently of being a normal or reduction cell. 
In \autoref{fig:NB301_representation} we visualize the adapted node attribute matrix and the adapted adjacency matrix to an exemplary DAG in the DARTS search space \cite{2018DARTS}. This is similar to the representation in \cite{2020Arch2vec}.

\subsection{NAS-Bench-NLP}\label{sec:supp_rep_nbnlp}

\begin{figure*}
	\centering
	\includegraphics[height=7cm]{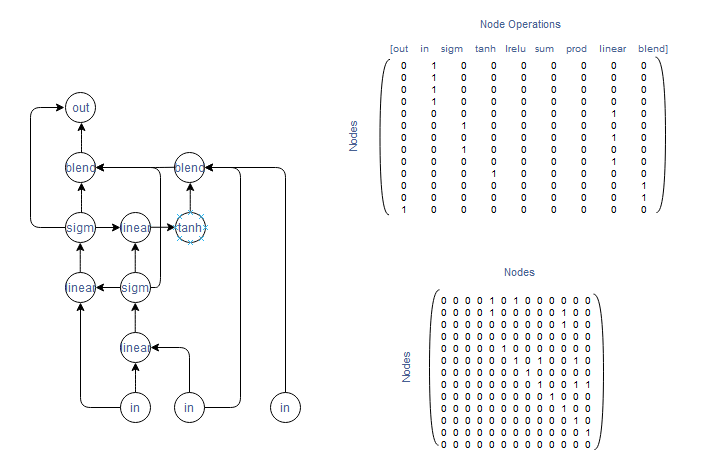}
	\caption{Exemplary cell representation from the NAS-Bench-NLP search space. (\textbf{left}) DAG representation of a graph with 12 nodes. (\textbf{right}) The top part shows the node attribute matrix to the DAG and the bottom part shows its adjacency matrix. \label{fig:NBNLP_representation}}
\end{figure*}
NAS-Bench-NLP \cite{2020NBNLP} is the first RNN-derived benchmark for language modeling tasks. From the total $10^{53}$ possible architectures in the complete search space, $14,322$ architectures are trained on Penn TreeBank \cite{2010PTB} (PTB) and provided in this benchmark. The cell search space is constrained by the number of nodes $\vert V \vert \leq 24$, the number of hidden states $\vert H \vert \leq 3$ and the number of linear input vectors $\leq 3$.  The nodes represent the architecture operational layer and are chosen from the set $\mathcal{O} = \{\textrm{linear}, \textrm{~element wise blending}, \textrm{~element wise product}, \\ \textrm{~element wise sum},  \textrm{~Tanh activation}, \textrm{~Sigmoid activation}, \textrm{~LeakyReLU activation}\}$.

For the experiments on NAS-Bench-NLP \cite{2020NBNLP} we make use of the surrogate benchmark NAS-Bench-X11 \cite{2021NBX11} and the additional  implementation in NAS-Bench-Suite \cite{NBSuite}. Note, for the NAS-Bench-X11 evaluations, each architecture from the NAS-Bench-NLP search space must be trained for three epochs to use the surrogate model, whereas NAS-Bench-Suite provides the surrogate model for NAS-Bench-NLP without learning curve information, but also accompanying a lower Kendall Tau rank correlation. For fast evaluations we use the latter surrogate for our experiments. In order to use the surrogate benchmark, the architecture representation is the same used in  \cite{2021NBX11} with the modification that each hidden node is connected to the output node. An exemplary architecture representation is visualized in \autoref{fig:NBNLP_representation}. A next step is to analyse the $14,332$ provided architectures on uniqueness, which leads to  $12,107$ unique architectures. Furthermore, since \cite{2021NBX11} and \cite{NBSuite} only provide a surrogate model, which only considers architectures with up to 12 nodes, we also restrict our training data to this subset leading to a total of $7,258$ architectures. 

We show experiments in the DARTs search space and on NAS-Bench-NLP in \autoref{sec:surr_benchmark_experiments}.

\subsection{Hardware-Aware-NAS-Bench}
The recently introduced HW-NAS-Bench \cite{2021HWNNB} is the first public dataset for hardware NAS.
It extends two representative NAS search spaces, NAS-Bench-201 \cite{2020NB201} and FBNet \cite{2019FBNet}, by providing measured and estimated hardware costs (i.e. latency and/or energy) for each device for all architectures in both search spaces.
For this, HW-NAS-Bench considers six hardware devices:
\textit{Edge~GPU} \cite{edgegpu}, \textit{Raspi~4} \cite{rapsi4}, \textit{Edge~TPU} \cite{edgetpu}, \textit{Pixel~3} \cite{pixel3}, \textit{ASIC-Eyeriss} \cite{eyeriss} and \textit{FPGA} \cite{fpga_1,fpga_2}.

In our experiments in \autoref{sec:hwbench-experiments} we consider the latency information on the NAS-Bench-201 search space.

\section{Additional Ablation Studies}
\label{supp:sec_ablation}

In this section we give an overview of different ablation studies with respect to the proposed AG-Net.
\subsection{Oracle Ablation}\label{supp:sec_push}
\begin{figure}
	\centering
	\includegraphics[height=3cm]{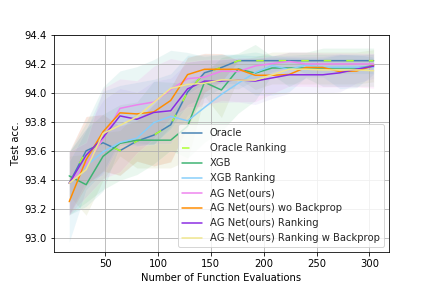}
	\caption{Architecture search on NAS-Bench-101. Reported is the mean over 10 trials for the search of the best architecture in terms of validation accuracy on the CIFAR-10 image classification task compared to strong predictor models.
		\label{fig:NB101_excellent_search}}
\end{figure}
As we have seen in the previous section, our model AG-Net is able to find high-scoring architectures in various search spaces of different sizes and with different objectives.
In addition, including the supposedly stronger predictor XGB \cite{XGB} leads to improvements for the search on NAS-Bench-NLP \cite{2020NBNLP}.
In this section, we include an even stronger architecture accuracy evaluation model, i.e. the benchmark query input itself (oracle).

The comparison of the oracle benchmark (also including the ranking metric as for XGB in the main paper) to our AG-Net and XGB modifications are visualized in \autoref{fig:NB101_excellent_search}.
This figure demonstrates the high performance of our model in the low query area.
The more queries are evaluated for the search, the better the oracle becomes, outperforming all other methods after $150$ queries.

\subsection{Latent Space Ablations}\label{supp:sec_baseline_ablation}
\begin{table*}[t]
	\caption{Ablation: Search results on NAS-Bench-101 and NAS-Bench-201 on the AG-Net latent space (mean over 10 trials with a maximal query amount of $300$).}
	\label{tab:NB201_Search_ablation}
	\scriptsize
	\begin{center}
		\resizebox{\textwidth}{!}{
			\begin{tabular}{c||c|c||c|c|c|c|c|c}
				\toprule
				& \multicolumn{2}{c||}{\textbf{NAS-Bench-101}} & \multicolumn{6}{c}{\textbf{NAS-Bench-201}}\\
				&  \multicolumn{2}{c||}{CIFAR-10}& \multicolumn{2}{c|}{CIFAR-10} & \multicolumn{2}{c|}{CIFAR-100} & \multicolumn{2}{c}{ImageNet16-120}    \\
				& Val. Acc  & Test Acc & Val. Acc  & Test Acc &Val. Acc & Test Acc &Val. Acc & Test Acc \\
				\midrule
				\textbf{Optimum}* &$95.06$ & $94.32$ & $91.61$ & $94.37$&  $73.49$ & $73.51$& $46.77$ & $47.31$ \\
				\midrule
				Random Search &  $94.27$ & $93.65$& $91.37$ & $ 93.92$ & $72.55$ & $72.49$ & $46.09$ & $46.05$ \\
				Local Search & $94.31$  & $93.66$ & $91.28$ & $ 94.01$ & $72.52$ & $72.59$ & $45.89$ & $46.07$  \\
				Bayesian Optimization & $94.27$  & $93.62$ & $91.30$ & $ 93.99$ & $72.23$ & $72.35$ & $46.09$ & $46.01$ \\
				Random Search + LSO& $94.64 $  & $\textbf{94.20}$ & $\textbf{91.61*}$ & $\textbf{94.37*}$ & $\textbf{73.49*}$ &$\textbf{73.51*}$ & $\textbf{46.77*}$ & $45.47$  \\
				Local Search + LSO &   $94.17$ & $93.50$ & $91.30$ & $93.96$ & $72.43$ & $72.58$ & $45.83$ & $45.95$  \\
				Bayesian Optimization +LSO & $94.50$ & $93.96$ & $91.43$ & $94.17$ & $72.64$  & $72.67$ & $46.30$ & $45.91$\\
				SGNAS \cite{2021SGNAS} + LSO &-& -& $\textbf{91.61*}$ & $ \textbf{94.37*}$ & $73.04$ & $73.12$ & $46.56$ & $\textbf{46.32}$\\ 
				
				AG-Net (ours) & $\textbf{94.96}$ & $\textbf{94.20}$& $\textbf{91.61*}$ & $ \textbf{94.37*}$ & $\textbf{73.49*}$ & $\textbf{73.51*}$ & $46.67$ & $46.22$  \\
				\bottomrule
		\end{tabular}}
	\end{center}
\end{table*}
\begin{figure*}[t]
	\centering
	\includegraphics[width=0.4\textwidth]{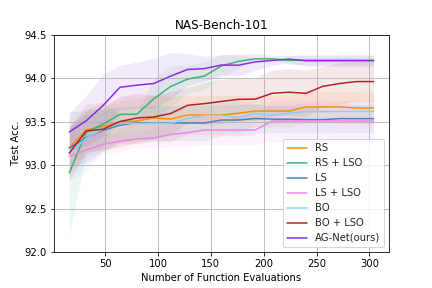}
	\includegraphics[width=0.4\textwidth]{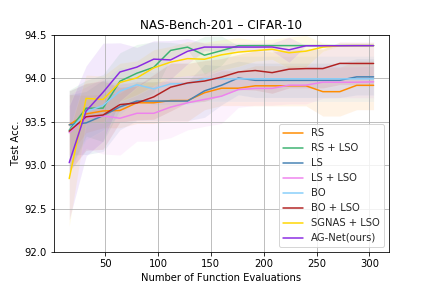}\\
	\includegraphics[width=0.4\textwidth]{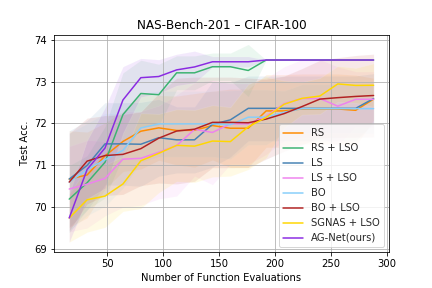}
	\includegraphics[width=0.4\textwidth]{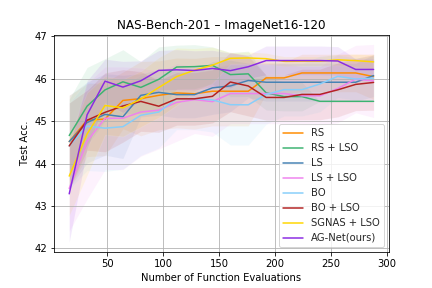}
	\caption{Ablation: neural architecture search on NAS-Bench-101 and  NAS-Bench-201 over 10 trials. 
		\label{fig:NB201_search_ablation} \label{fig:NB101_search_ablation}}
\end{figure*}

As we have seen in \autoref{sec:tabluar_benchmark_experiments}, AG-Net improves over state-of-the art methods.
For additional comparisons, we investigate different search methods in the latent space of the generative model, with samples $z$ from a grid and also include baselines using the LSO approach. For the first experiment we use the generator solely as a data sampler from the generator's latent space without any retraining, for the latter baseline we retrain the generator during the search.
For the optimization, we use Bayesian optimization, local search and random search. 

\paragraph{Bayesian Optimization}
We use DNGO\cite{2015DNGO} as our uncertainty prediction model for the Bayesian optimization search strategy, with the basis regression network being a one-layer MLP with a hidden dimensionality of 128, which is trained for 100 epochs and expected improvement (EI) \cite{74Mockus} as our acquisition function, which is mostly used in NAS. We set the best function value for the EI evaluation as the best validation accuracy of the training data.
We sample $16$ initial random latent space variables $\textbf{z} \sim \mathcal{U}[-3,3]$ and decode them to graph data using our pretrained generative model.
These latent space variables and their corresponding validation architecture performances are then the inputs for the DNGO model for training.
Again, the best 16 architectures are selected using EI in each round to be evaluated and added to the training data.
This search ends when the total query amount of $300$ is reached.

\paragraph{Random and Local Search}
In addition to Bayesian Optimization as a comparison, we also include a random search \cite{2019RS} and local search investigation. 
Recently, \cite{2020LocalSearchNAS} show that local search is a powerful NAS baseline, resulting in competitive results.
Local search \cite{2020LocalSearchNAS} evaluates samples and their neighborhood uniformly at random.
An option to define the neighborhood is the set of architectures which differ from a sampled architecture by one node or edge.
This can be done only in the discrete search space, given for example by the tabular NAS-Benchmarks.
We have to adapt the neighborhood definition in our latent space for local search in this space.
We sample a latent space variable $\textbf{z} \sim \mathcal{U}[-3,3]$, decode it and evaluate the generated neural architecture.
Here, we define neighborhood as the Euclidean space around the sampled latent variable $ U_{\epsilon}(z) = \{y \sim \mathcal{U}[-3,3] \vert d(z,y) < \epsilon\}$, with $\epsilon $ being sufficiently small.
This neighborhood is then investigated until a local optimum in terms of validation accuracy is reached.
Furthermore, we include a random search and local search comparison using weighted retraining.
Here, we retrain the generative model in each search iteration for $1$ epoch with the weighted objective function, ceteris paribus.

To compare with weight-sharing approaches, we also compare to the supernet from \cite{2021SGNAS} for the NAS-Bench-201 search space.
To compare our AG-Net with SGNAS, we use the supernet as our surrogate model to predict the architectures performance while retraining the generative model in the weighted manner. 
The results of our ablation studies are reported in \autoref{tab:NB201_Search_ablation}.
AG-Net improves over search methods on the latent space with and without LSO on both benchmarks, demonstrating that our generator in combination with our MLP surrogate model learns to adapt the distribution shift constructed by the weighted retraining best.

For further visualizations we also plot different ablation search methods over different query numbers in \autoref{fig:NB201_search_ablation} for both benchmarks NAS-Bench-101 and NAS-Bench-201.
This figure demonstrates the high any-time performance of our method on both search spaces.
For any number of available queries, our model is better in finding high-performing architectures from the latent space than other latent space based methods. 

\subsection{Predictor Ablation -- Local Solution}
\begin{figure}
	\centering
	\includegraphics[height=3cm]{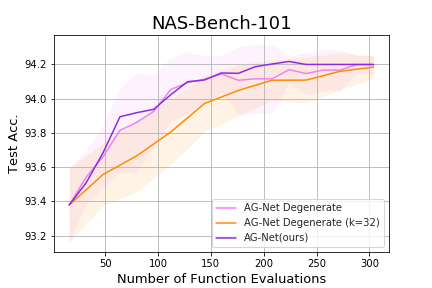}
	\caption{Architecture search on NAS-Bench-101 in the degenerate setting. Reported is the mean over 10 trials.
		\label{fig:local_optima}}
\end{figure}
Our proposed method, consisting of the generative and surrogate model combined with the latent space optimization, makes the architecture search focus on promising regions in the search space. This method could be trapped in local solutions, which we investigate experimentally in the following. First, the previous section already points out that our proposed method AG-Net improves over both local search methods with and without the latent space optimization approach. Thus, we assume that the latent space optimization learns properties of high-scoring architectures without being easily trapped in poor local solutions. The amount of samples drawn in each search iteration also provides a trade-off between diversity versus specificity. To investigate further how easily AG-Net could be trapped in a local solution, we test our method when it only uses in total the best k (predicted) architectures from our test samples and the training data as a new training set for the next search iteration (degenerative) and is thereby encouraged to forget about worse performing architectures. \autoref{fig:local_optima} shows the search behaviour of the degenerative model with $k=16$ and $k=32$. Even in this case, AG-Net is not easily trapped in poor solutions.

\section{Experiments: Implementation Details}
\label{sup:implementation-details}

\subsection{Surrogate Model}
In this section, we present details about the surrogate models used in the main paper.
The MLP surrogate model used for our AG-Net is a 4-layer MLP with ReLU non-activation functions.
The hidden size equals the input size.
The input to the MLP surrogate model is the vector representation $\in \mathbb{R}^n$ of our graphs: a concatenation of the flatted node attribute matrix and flatted upper triangular matrix of the adjacency matrix, which presents the edge scores, see \autoref{sup:sp-representations} for visualizations.
Note, the vector dimension $n$ differs across the search spaces due to the different maximal amount of nodes.
Our AG-Net passes the output of our generator, i.e. a generated vector representation, as the direct input to our MLP surrogate model.

We consider as an alternative surrogate model the XGB \cite{XGB} prediction model.
The input to this prediction model is the vector representation of the architecture.

\subsection{Search Algorithm}

High-level descriptions of the unconstrained (\autoref{sec:tabluar_benchmark_experiments}) and constrained (\autoref{sec:hwbench-experiments}) versions of our search algorithm are depicted in \autoref{alg:nas-unconstrained} and \autoref{alg:nas-constrained} respectively.

\begin{algorithm}[ht]
	\SetAlgoLined
	\caption{Unconstrained Search Algorithm}
	\label{alg:nas-unconstrained}
	\KwInput{(i) Search space $p_D$}
	\KwInput{(ii) Pretrained generator $G$}
	\KwInput{(iii) Untrained performance predictor $P$}
	\KwInput{(iv) Query budget $b$}
	\KwInput{(v) $e$ epochs to train $G$ and $P$}
	\Comment{Initialize training data}
	$\mathbf{D} \leftarrow \{\}$ \\
	\While{$\vert \mathbf{D} \vert < 16$}{
		$\mathbf{D} \leftarrow \mathbf{D} \cup \{d\sim p_D\}$ \\
	}
	\Comment{Evaluate architectures (get accuracies on target image dataset)}
	$\mathbf{D} \leftarrow \text{eval}(\mathbf{D})$ \\
	\Comment{Randomly initialize predictor weights}
	${P} \leftarrow \text{init}(P)$ \\
	\Comment{Search loop}
	\While{$\vert \mathbf{D} \vert < b$}{
		\Comment{Weight training data by performance}
		$\mathbf{D}_{\text{w}} \leftarrow  \text{weight}(\mathbf{D})$ \\
		\Comment{Train generator and predictor}
		train($G,P$, $\mathbf{D}_{\text{w}}$, $e$) \\
		\Comment{Generate 100 candidates}
		$\mathbf{D}_{\text{cand}} \leftarrow \{\}$ \\
		\While{$\vert \mathbf{D}_{\text{cand}} \vert < 100$}{
			$\mathbf{z} \sim \mathcal{U}[-3,3]$ \\ 
			$\mathbf{D}_{\text{cand}} \leftarrow \mathbf{D}_{\text{cand}} \cup G(\mathbf{z})$ \\
		}
		\Comment{Select top 16 candidates with P}
		$\mathbf{D}_{\text{cand}} \leftarrow \text{select}(\mathbf{D}_{\text{cand}},P,$16$)$ \\
		\Comment{Evaluate and add to data}
		$\mathbf{D} \leftarrow \mathbf{D} \cup \text{eval}(\mathbf{D}_{\text{cand}})$
	}
\end{algorithm}

\begin{algorithm}[ht!]
	\SetAlgoLined
	\caption{Constrained Search Algorithm}
	\label{alg:nas-constrained}
	\KwInput{(i) Search space $p_D$}
	\KwInput{(ii) Pretrained generator $G$}
	\KwInput{(iii) Untrained performance predictor $P_a$}
	\KwInput{(iv) Set of constraint predictors $P_c$ }
	\KwInput{(v) Query budget $b$}
	\KwInput{(vi) $e$ epochs to train $G$ and $P$}
	\KwInput{(vii) Set of constraints $C$}
	\Comment{Initialize training data}
	$\mathbf{D} \leftarrow \{\}$ \\
	\While{$\vert \mathbf{D} \vert < 16$}{
		$\mathbf{D} \leftarrow \mathbf{D} \cup \{d\sim p_D\}$ \\
	}
	\Comment{Evaluate architectures (get accuracies and constraints on target image dataset)}
	$\mathbf{D} \leftarrow \text{eval}(\mathbf{D})$ \\
	\Comment{Randomly initialize predictor weights}
	${P_a} \leftarrow \text{init}(P_a)$ \\
	\ForEach{$P \in P_c$}{
		${P} \leftarrow \text{init}(P)$
	}
	\Comment{Search loop}
	\While{$\vert \mathbf{D} \vert < b$}{
		\Comment{Weight train data by performance and constraints}
		$\mathbf{D}_{\text{w}} \leftarrow  \text{weight}(\mathbf{D}, C)$ \\
		\Comment{Train generator and predictors}
		train($G,P_a, P_c$, $\mathbf{D}_{\text{w}}$, $e$) \\
		\Comment{Generate 100 candidates}
		$\mathbf{D}_{\text{cand}} \leftarrow \{\}$ \\
		\While{$\vert \mathbf{D}_{\text{cand}} \vert < 100$}{
			$\mathbf{z} \sim \mathcal{U}[-3,3]$ \\
			$\mathbf{D}_{\text{cand}} \leftarrow \mathbf{D}_{\text{cand}} \cup G(\mathbf{z})$ \\
		}
		\Comment{Select top16 candidates with ${P_{a}}$ and $P_c$}
		$\mathbf{D}_{\text{cand}} \leftarrow \text{select}(\mathbf{D}_{\text{cand}},P_a,P_c,$16$)$ \\
		\Comment{Evaluate and add to data}
		$\mathbf{D} \leftarrow \mathbf{D} \cup \text{eval}(\mathbf{D}_{\text{cand}})$
	}
\end{algorithm}

\subsection{NAS-Bench-101}

\begin{table}[h]
	\scriptsize
	\begin{center}
		\caption{Architecture search on NAS-Bench-101. Reported is the mean and the standard deviation over 10 trials for the search of the best architecture in terms of validation accuracy on the CIFAR-10 image classification task compared to state-of-the-art methods.}
		\label{tab:NB101_Search_supp}
		\begin{tabular}{c||c|c|c|c||c}
			\toprule
			\textbf{NAS Method}  & \textbf{Val. Acc} ($\%$) & \textbf{StD} ($\%$)& \textbf{Test Acc} ($\%$) &  \textbf{StD} & ($\%$)\textbf{Queries} \\
			\midrule
			\textbf{Optimum*} & $95.06$ & - & $94.32$& - & \\
			\midrule
			Arch2vec + RL \cite{2020Arch2vec} & - & - & $94.10$ & - & $400$\\
			Arch2vec + BO \cite{2020Arch2vec}& - & - & $94.05$ & - & $400$\\
			NAO \textsuperscript{\ddag}\cite{2018NAO} & $94.66$ & $0.14 $ &
			$93.49 $ & $0.59 $ & 192 \\
			BANANAS\textsuperscript{\textdagger} \cite{2021BANANAS} & $94.73 $ & $ 0.17$ & $94.09 $ & $0.19$ & 192 \\
			Bayesian Optimization\textsuperscript{\textdagger} \cite{2015DNGO}  & $94.57 $ & $ 0.2 $ &  $93.96 $ & $ 0.21$ & 192 \\
			Local Search\textsuperscript{\textdagger} \cite{2020LocalSearchNAS}  & $94.57 $ & $ 0.15$ & $93.97 $ & $ 0.13$ & 192\\
			Random Search\textsuperscript{\textdagger}\cite{2019RS}  & $94.31 $ & $ 0.15$ & $93.61 $ & $ 0.27$  & 192 \\
			Regularized Evolution\textsuperscript{*}\cite{2019EvolutionaryNAS}  & $94.47 $ & $ 0.11$ & $93.89 $ & $ 0.2$ & 192 \\
			WeakNAS \cite{2021WeakNAS} & - & - & $\textbf{94.18} $ & $ 0.14$ & 200 \\
			\midrule
			XGB (ours) & $94.61 $ & $ 0.04$ & $94.13 $ & $ 0.11 $ & 192 \\
			XGB + ranking (ours) & $94.60 $ & $ 0.08$ & $94.14 $ & $ 0.19$ & 192 \\
			AG-Net (ours)  & $\textbf{94.90} $ & $ 0.22$ & $\textbf{94.18} $ & $ 0.10$  & 192
			\\
			\bottomrule
		\end{tabular}
	\end{center}
\end{table}

In this section, we give more information about the NAS-Bench-101 experiments from the main paper. 

\autoref{tab:NB101_Search_supp} is the detailed version of \autoref{tab:NB101_Search} including the standard deviation.

\subsection{NAS-Bench-201}

\autoref{tab:NB201_Search_supp} is the detailed version of \autoref{tab:NB201_Search} including the standard deviation.

\begin{table*}[ht]
	\scriptsize
	\caption{Architecture Search on NAS-Bench-201. We report the mean and standard deviation over 10 trials for the search of the architecture with the highest validation accuracy.
		For comparable numbers of queries, AG-Net performs similarly or better than the previous state of the art.}
	\label{tab:NB201_Search_supp}
	\begin{center}
		\resizebox{\textwidth}{!}{
			\begin{tabular}{c||c|c|c|c||c|c|c|c||c|c|c|c||c}
				\toprule
				\textbf{NAS Method}  & \multicolumn{4}{c||}{\textbf{CIFAR-10}} & \multicolumn{4}{c||}{\textbf{CIFAR-100}} & \multicolumn{4}{c||}{\textbf{ImageNet16-120}}  & \textbf{Queries} \\
				& Val. Acc & StD & Test Acc & StD &Val. Acc & StD & Test Acc & StD  &Val. Acc & StD & Test Acc & StD \\
				\midrule
				\textbf{Optimum*} & $91.61$ & & $94.37$&  &$73.49$& & $73.51$&& $46.73$& & $47.31$ & & \\
				\midrule
				SGNAS  \cite{2021SGNAS} & $90.18 $&$ 0.31$ & $93.53 $&$ 0.12$ &  $70.28 $&$ 1.2$ & $70.31 $&$ 1.09$ &  $44.65 $&$ 2.32$ & $44.98 $&$ 2.10$ & \\
				\midrule
				Arch2vec + BO \cite{2020Arch2vec}  &  $91.41 $&$ 0.22$  & $94.18 $&$ 0.24$ &  $\textbf{73.35} $&$ 0.32$ & $\textbf{73.37} $&$ 0.30$  &  $46.34 $&$ 0.18$ & $46.27 $&$ 0.37$  & 100  \\
				AG-Net (ours)& $\textbf{91.55} $&$ 0.08$ & $\textbf{94.24} $&$ 0.19$ & $73.2 $&$ 0.34$ & $73.12 $&$ 0.40$ & $46.31 $&$ 0.33$ & $46.2 $&$ 0.47$ &96 \\
				AG-Net (ours with topk=1) & $91.41 $&$ 0.30$ & $94.16$&$ 0.31$ & $73.14 $&$ 0.56$ & $73.15 $&$ 0.54$ & $\textbf{46.42} $&$ 0.14$ & $\textbf{46.43} $&$ 0.30$ &100\\
				\midrule
				BANANAS\textsuperscript{\textdagger} \cite{2021BANANAS}  & $91.56 $&$ 0.14$ & $94.3 $&$0.22$ & $\textbf{73.49*} $&$ 0.00$ & $73.50 $&$ 0.00$ & $\textbf{46.65} $&$ 0.13$& $\textbf{46.51} $&$ 0.11 $ & 192 \\
				BO\textsuperscript{\textdagger}  \cite{2015DNGO}  & $91.54$&$ 0.06$  & $94.22$&$ 0.18$ & $73.26 $&$ 0.19$ & $73.22 $&$ 0.27$ &  $46.43 $&$ 0.35$  & $46.40 $&$ 0.35$ & 192 \\
				RS \textsuperscript{\textdagger} \cite{2019RS}   & $91.12 $&$ 0.26$ & $93.89 $&$ 0.27$ & $72.08 $&$ 0.53$ & $72.07 $&$ 0.61$ & $45.87 $&$ 0.39$ & $45.98 $&$ 0.41$ & 192  \\
				XGB (ours) & $91.54 $&$ 0.09$ & $94.34 $&$ 0.10 $ & $73.10 $&$ 0.51 $ & $72.93 $&$ 0.74 $ &  $46.48 $&$ 0.13$  & $46.08 $&$ 0.79$ & 192  \\
				XGB + Ranking (ours) & $91.48 $&$ 0.12 $ & $94.25 $&$ 0.15$ & $73.20  $&$ 0.36$ & $73.24  $&$ 0.34$ & $46.40 $&$ 0.28$ & $46.16 $&$ 0.64$ & 192 \\
				AG-Net (ours) & $\textbf{91.60} $&$ 0.02 $ & $\textbf{94.37*} $&$ 0.00$ & $\textbf{73.49*} $&$ 0.00$ & $\textbf{73.51*} $&$ 0.00$ & $46.64 $&$ 0.12$ & $46.43 $&$ 0.34$ &192  \\
				\midrule
				GANAS \cite{2021GANAS}& - & - & $94.34$&$ 0.05$  &- & - & $73.28 $&$ 0.17$ & - &- & $\textbf{46.80} $&$ 0.29$ & 444  \\
				AG-Net (ours) & $\textbf{91.61*} $&$ 0.00$ & $\textbf{94.37*} $&$ 0.00$   &  $\textbf{73.49*} $&$ 0.00$ & $\textbf{73.51*} $&$ 0.00$ &  $\textbf{46.73*} $&$ 0.00$ & $46.42 $&$ 0.00$ & 400  \\
				\bottomrule
		\end{tabular}}
	\end{center}
\end{table*}

\subsection{DARTS Search Space}\label{sec:supp_darts}
\paragraph{Additional Results}
\autoref{supp_tab:NB301_Search} is the detailed version of \autoref{tab:NB301_Search} including the standard deviation.
\begin{table}[ht]
	\caption{Results on NAS-Bench-301 (mean and standard deviation over 50 trials) for the search of the best architecture in terms of validation accuracy compared to state-of-the-art methods.}
	\label{supp_tab:NB301_Search}
	\scriptsize
	\begin{center}
		\begin{tabular}{c||c| c || c}
			\toprule
			\textbf{NAS Method}  & \textbf{Val. Acc} ($\%$) & \textbf{StD} ($\%$) & \textbf{Queries} \\
			\midrule
			BANANAS\textsuperscript{\textdagger} \cite{2021BANANAS} & $94.77$  & $0.10$  & 192 \\
			Bayesian Optimization\textsuperscript{\textdagger}  \cite{2015DNGO}  & $94.71$ & $0.10$ & 192  \\
			Local Search\textsuperscript{\textdagger} \cite{2020LocalSearchNAS} & $\textbf{95.02}$ & $0.10$ & 192\\
			Random Search\textsuperscript{\textdagger}\cite{2019RS}  & $94.31$ & $0.12$ & 192 \\
			Regularized Evolution\textsuperscript{{\textdagger}}\cite{2019EvolutionaryNAS}  & $94.75$ &  $0.11$& 192 \\
			\midrule
			XGB (ours) & $94.79$ & $0.13$ & 192 \\
			XGR + Ranking (ours) & $94.76$ & $0.14$& 192 \\
			\midrule
			AG-Net (ours)  & $94.79$ & $0.12$ & 192
			\\
			\bottomrule
		\end{tabular}
	\end{center}
\end{table}

\label{sec:darts}
\paragraph{Search Process using NAS-Bench-301}

For experiments in the DARTS \cite{2018DARTS} search space, we first train our generative model on generating valid cells, as visualized in \autoref{fig:NB301_representation}; here we do not distinguish between generating a normal or a reduction cell. Having a pretrained generative model for generating valid cell representations in the DARTS search space allows for searching well-performing architectures.
Here we describe the search process for architectures evaluated on CIFAR-10 using the surrogate benchmark NAS-Bench-301 \cite{2020NB301}. Since the DARTS search space is defined by a normal and reduction cell, we have to adapt the search process, compared to the search in the tabular benchmark search spaces, where the architectures differ between the DAG. We begin the search by randomly sampling $16$ architectures from NAS-Bench-301. Next, we generate one normal cell.
This cell is used to search for the best reduction cell in terms of the accuracy given by the surrogate benchmark NAS-Bench-301, in combination with the randomly sampled cell.
This search procedure then follows the same steps as for the tabular benchmarks and stops after we reach a query amount of 155.
Now, we can use the best found reduction cell as a fixed starting point to search for the best normal cell in the same manner as before.
The overall search stops after a maximal amount of 310 queries.
The search outcome differs between starting with a reduction or the normal cell.
The search procedure starting with a random reduction cell is analogous.
In the main paper, we report the search outcome for NAS-Bench-301 \cite{2020NB301} starting with a random reduction cell.

\paragraph{Search Process using TENAS}
As we described in the previous section, the search in the DARTS \cite{2018DARTS} search space needs adaptions in the search procedure.
Here we describe the further adaption of using training free measurements instead of the NAS-Bench-301 prediction.
The training free measurements are based on the recent paper TE-NAS \cite{2021Im4GPU}, which ranks architectures by analysing the neural tangent kernel, by its condition number (KN), and the number of linear regions (NLR) of each architecture. 
Concretely, for the search on ImageNet \cite{2009ImageNet} we search for architectures in terms of their KN value and their number of linear regions instead of their validation accuracy.
In the beginning of our search we generate three random normal cells.
These cells are used to search for an optimal reduction cell optimizing both KN and NLR measurements.
In each search iteration we generate reduction cells and calculate the KN and NLR for each combination of normal cell and reduction cell.
The reduction cells are ranked according to their mean KN and their mean NLR (mean in terms of all three normal cells).
The 16 best ranked reduction cells are then used for the next iteration of reduction cell search.
The reduction cell search stops, when a maximum of 104 queries is reached.
After that we use the best found reduction cell in terms of the lowest KN and the highest NLR for the next search for a normal cell.
The next steps use this best found reduction cell as a starting point and searches for the best normal cell in the same manner as before. 
The search stops after a total of 208 queries and outputs an overall normal and reduction cell combination, leading to a DARTS \cite{2018DARTS} architecture, which we train on ImageNet \cite{2009ImageNet} using the same training pipeline as \cite{2021Im4GPU}.

\subsection{NAS-Bench-NLP}
\autoref{tab:NBNLP_Search_supp} is the detailed version of \autoref{tab:NB301_Search} including the standard deviation.
\begin{table}[t]
	\caption{Results on NAS-Bench-NLP (mean and standard deviation over 100 trials) for the search of the best architecture in terms of validation perplexity compared to state-of-the-art methods.}
	\label{tab:NBNLP_Search_supp}
	\scriptsize
	\begin{center}
		\begin{tabular}{c||c|c || c}
			\toprule
			\textbf{NAS Method}  & \textbf{Val. Perplexity} ($\%$) & \textbf{StD} ($\%$) & \textbf{Queries} \\
			\midrule
			BANANAS\textsuperscript{\textdagger} \cite{2021BANANAS} & $95.68$  & $0.16$  & 304 \\
			Local Search\textsuperscript{\textdagger} \cite{2020LocalSearchNAS} & $95.69$ & $0.18$ & 304\\
			Random Search\textsuperscript{\textdagger}\cite{2019RS}  & $95.64$ & $0.19$ & 304 \\
			Regularized Evolution\textsuperscript{{\textdagger}}\cite{2019EvolutionaryNAS}  & $95.66$ &  $0.21$& 304 \\
			\midrule
			XGB (ours) & $\textbf{95.95}$ & $0.20$ & 304 \\
			XGR + Ranking (ours) & $95.92$ & $0.19$& 304 \\
			\midrule
			AG-Net (ours)  & $95.86$ & $0.18$ & 304
			\\
			\bottomrule
		\end{tabular}
	\end{center}
\end{table}

\subsection{Hardware-Aware NAS-Bench}
In comparison to the experiments for NAS-Bench-101 \cite{2019NB101} and NAS-Bench-201 \cite{2020NB201} image benchmarks, the search on the Hardware-Aware NAS-Bench \cite{2021HWNNB} changes to be a multi-objective learning procedure.
We compare two different objective settings: i) a joint constrained optimization in \autoref{eq:joint=1_optimization}
and ii) a constrained optimization in \autoref{eq:joint=0_optimization}.
For both settings we need to adapt the surrogate model by including an additional predictor $g(\cdot)$ for latency.
We implement $g(\cdot)$ equally to the performance predictor $f(\cdot)$, whereas both predictors share weights in our experiments.
We give a detailed overview of the hyperparameter settings in \autoref{sec:HP}.
Since we include an additional predictor, the training objective needs to be updated, as seen in \autoref{eq:hw_cond_loss} with multiple targets.
The risk of including multiple targets to the training objective is an exploding loss leading to reduced valid generation ability of our generative network.
In order to overcome this problem, we scale each loss term by the largest one, such that each term is at most $1$.
This way, we have a more stable training.

\begin{figure}[ht]
	\centering
	\includegraphics[width=0.45\textwidth]{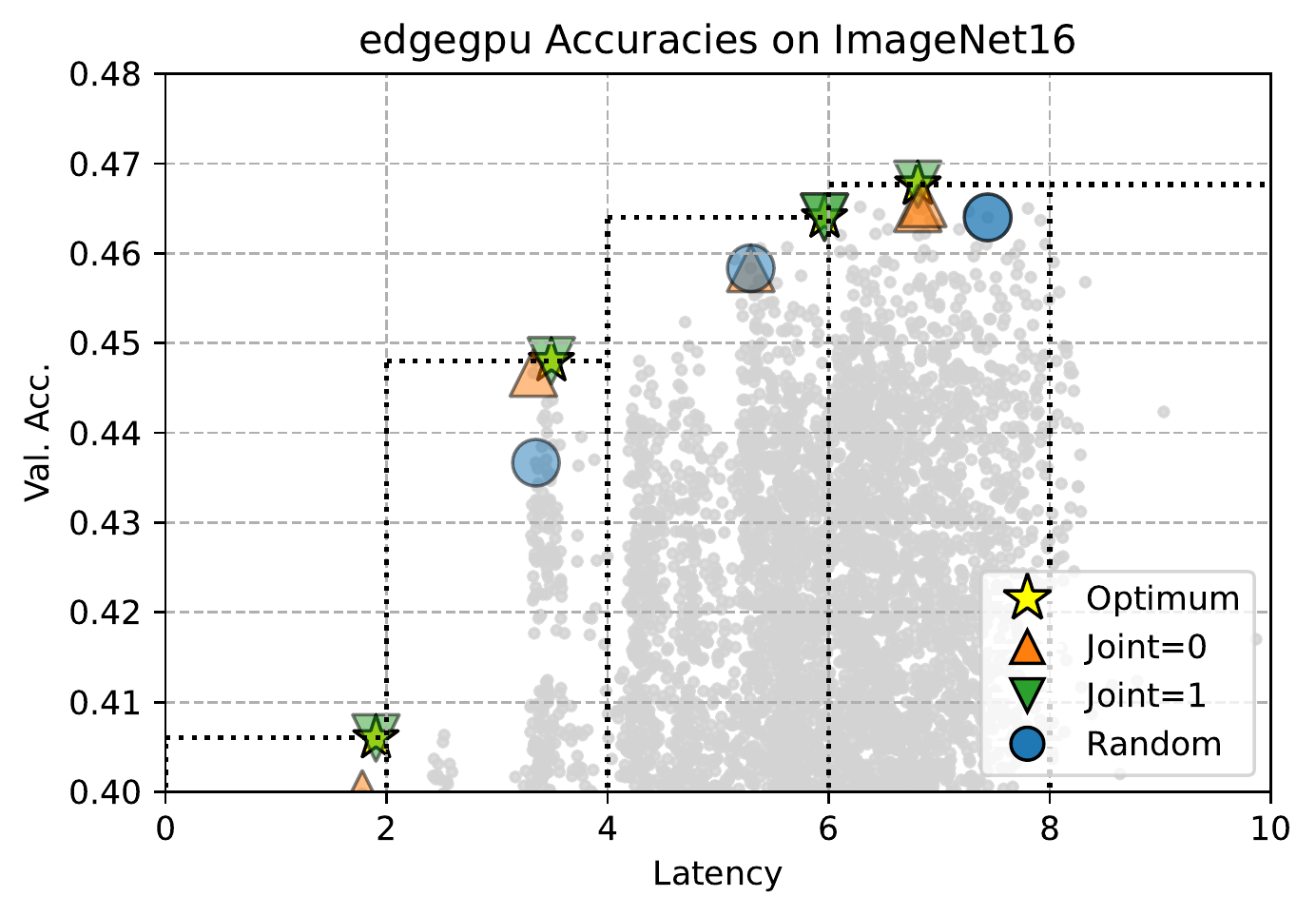}
	\includegraphics[width=0.45\textwidth]{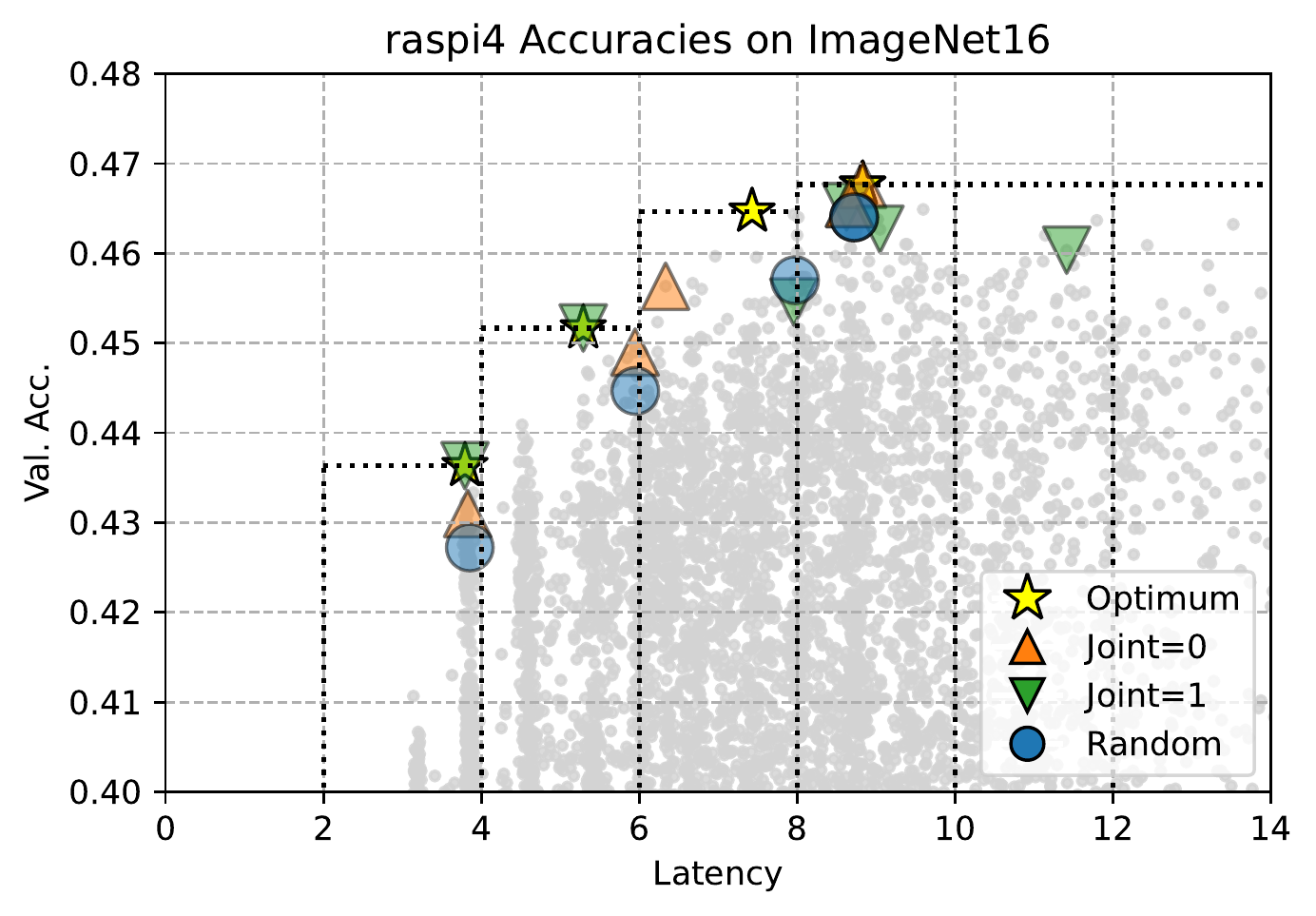}\\
	\includegraphics[width=0.45\textwidth]{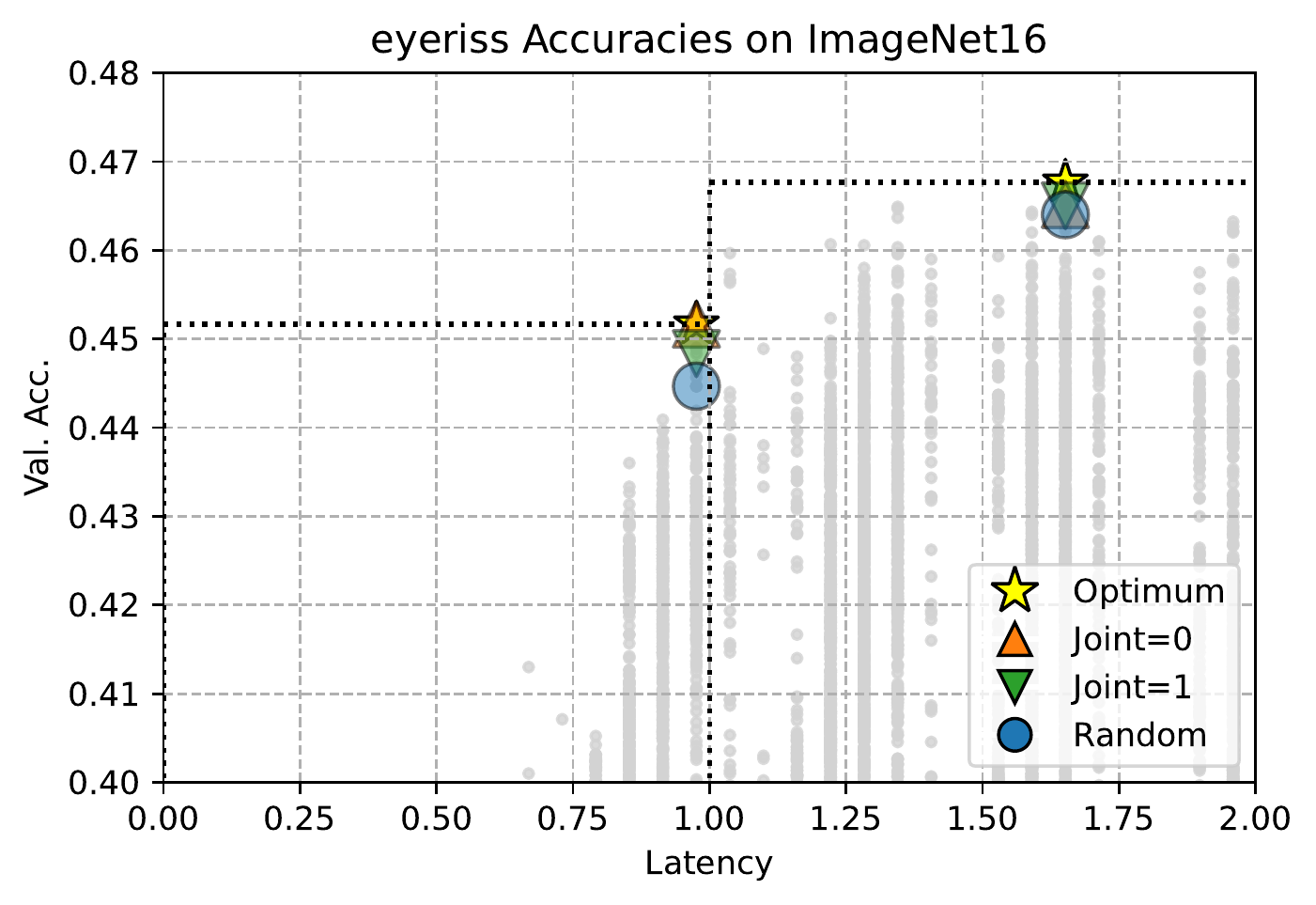}
	\includegraphics[width=0.45\textwidth]{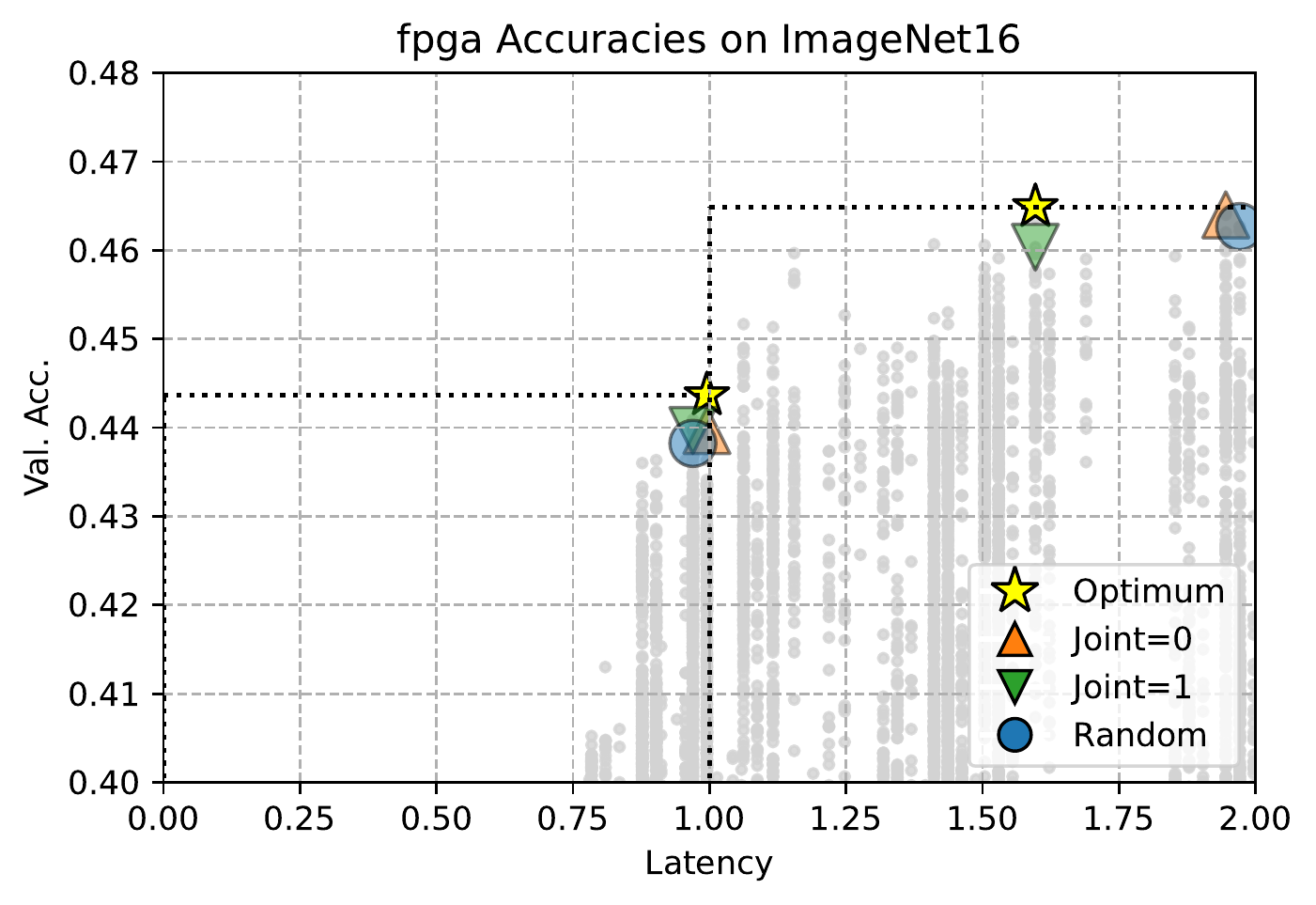}
	\caption{
		Exemplary searches on HW-NAS-Bench for image classification on ImageNet16 with $192$ queries on Edge GPU, Raspi4, Eyeriss, FPGA and latency conditions $L\in\{2,4,6,8,10\}$, $L\in\{2,4,6,8,10,12,14\}$ and $L\in\{1,2\}$ (y-axis zoomed for visibility).\label{fig:HW_search_edgegpu} \label{fig:HW_search_raspi4} \label{fig:HW_search_eyeriss} \label{fig:HW_search_fpga}}
\end{figure}

\paragraph{Exemplary Searches for Other Devices}
In \autoref{fig:HW_search} we showed an exemplary search result comparing random search with both of our constrained algorithm settings in the case of different latency constraints on a Pixel3.
In the following, we show more examples on different devices in \autoref{fig:HW_search_edgegpu}.
These plots show that both methods \textit{Joint=1} and \textit{Joint=0} outperform the random search baseline in all different device experiments. The same results as in the main paper holds therefore for all other devices too; \textit{Joint=1} is able to find better performing architectures compared to \textit{Joint=0} if the latency constraint $L$ restricts the feasible search space strongly.

\paragraph{Search Progress and Baselines}
\begin{figure}[h]
	\centering
	\includegraphics[width=0.45\linewidth]{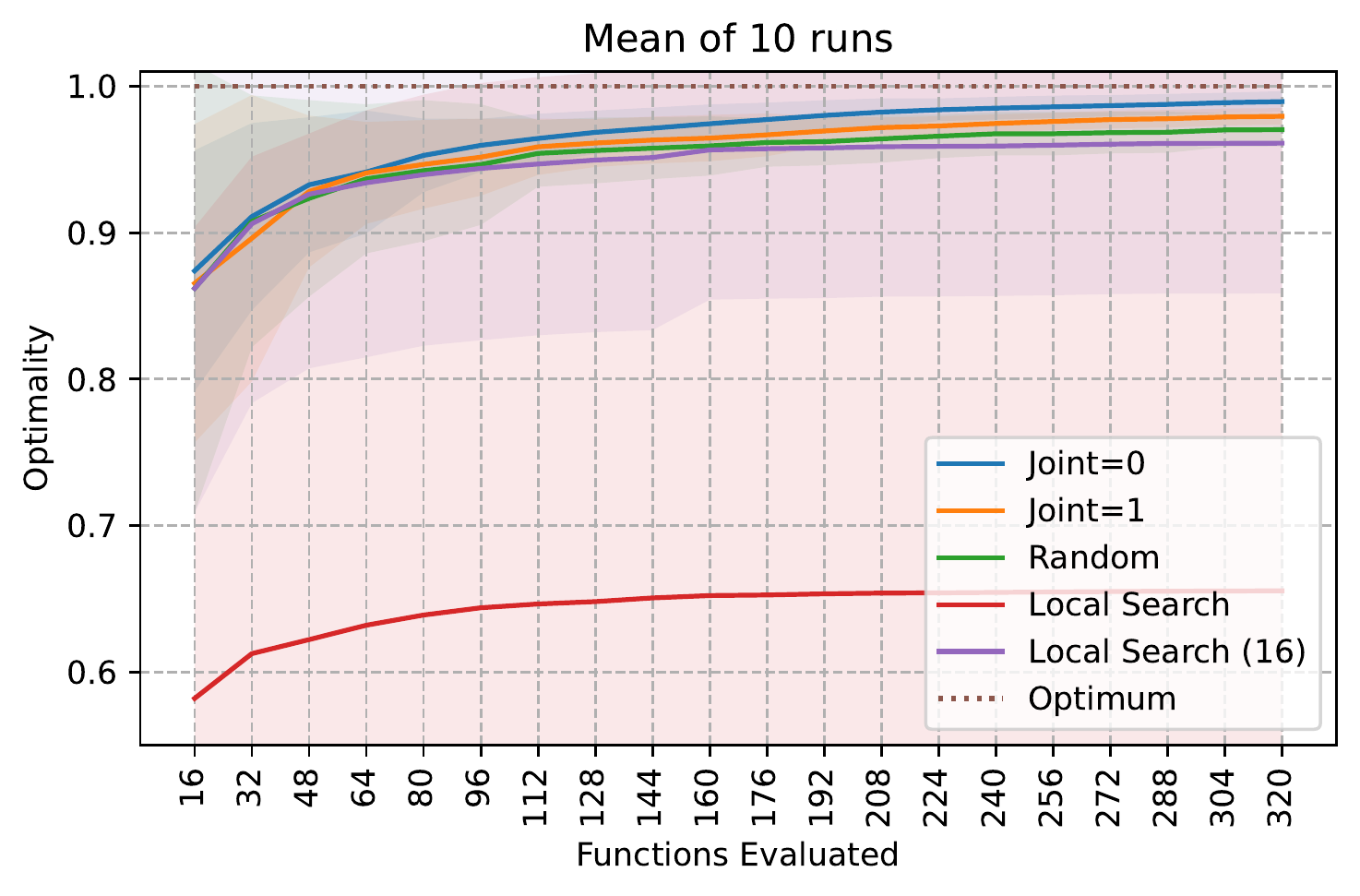}
	\includegraphics[width=0.45\linewidth]{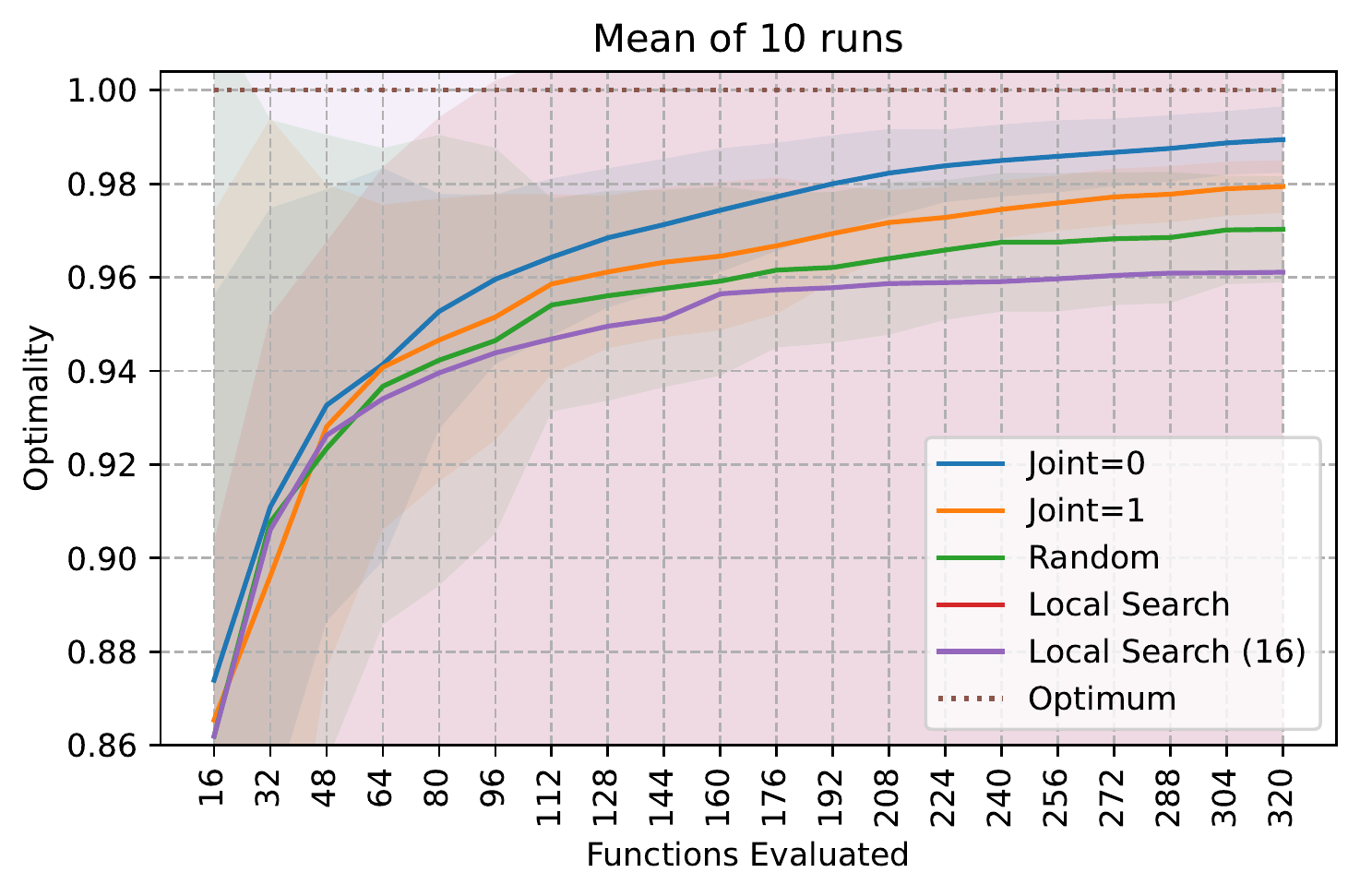}
	
	\caption{(\textbf{left})
		Optimality for all search parameters in \autoref{tab:hwbench-results} at any time during the search progress in terms of the number of evaluated architectures (up to $320$).
		Optimality is the mean validation accuracy of $10$ runs per algorithm, normalized by the optimal value for each parameter setting (hence, optimum is at $1.0$). (\textbf{right}) zoomed y-axis
	}
	\label{fig:hwnas-mean}\label{fig:hwnas-mean-zoom}
\end{figure}
Local search \cite{2020LocalSearchNAS} is considered a strong baseline in NAS.
In the case of constrained searches (as in HW-NAS-Bench), we noticed that it cannot perform well without adaptation.
The vanilla local search algorithm expects as input a single randomly drawn architecture from the search space.
However, this architecture is not guaranteed to be feasible in this setting, as its latency can be larger than the latency constraint.
To circumvent this, we performed local search in the following settings: (a) local search vanilla setting with one randomly drawn architecture, and (b) local search initialized with $16$ randomly drawn architectures.
In each setting, local search continues to search the neighborhood of the next best architecture in terms of accuracy that satisfies the latency constraint.
We noticed that initializing local search with $16$ randomly drawn architectures improves its performance substantially, however, it is still not on par with random search \cite{2019RS} in this constrained search space.
Consequently, we only show random search as the baseline in \autoref{tab:hwbench-results} to improve readability.
In \autoref{fig:hwnas-mean} we show the progress of our algorithms (\emph{Joint=0} and \emph{Joint=1}) compared to random search and local search in settings (a) and (b).

\section{Generator Details}\label{sec:generator}\label{supp:sec_generator_ability}
\subsection{Generator Evaluation}
Based on an investigation of autoencoder abilities from \cite{2020Arch2vec} and \cite{2021SVGe}, we can examine the generation ability of our generative model.
For that we train our generator on $90\%$ of the overall dataset, and thus have a hold-out dataset of $10\%$ for the tabular benchmarks. The generative model training on the surrogate benchmarks is a priori only on a subset of the overall dataset.
Additionally, we sample $10,000$ random variables $\textbf{z}\sim \mathcal{N}(0,1)$ and decode them to graphs.
We report the results of this investigation in \autoref{tab:Generator_Abilites}.
Here, validity describes the ratio of valid graphs our generator model generates,
uniqueness describes the portion of unique graphs from the valid generated graphs, and novelty is the portion of generated graphs \emph{not in the training set}. It is not surprising for the NAS-Bench-301 and NAS-Bench-NLP search spaces, that our model is able to generate $100 \%$ unique and novel graphs, given the large size of both search spaces.

This demonstrates that our simple generator model is able to generate valid graphs with high novelty and consequently is able to cover a substantial part of the search space.

\autoref{tab:Generator_Abilites} also reports the training costs of the generative model on the complete dataset as described in \autoref{sup:sp-representations} on a single Tesla V100.
We used for the experiments the OMNI cluster from the University of Siegen. 

\begin{table}[t]
	\caption{Generator Abilities and training costs. The proposed generator generates architectures with high validity and uniqueness scores. The novelty scores are in a similar range as for previous methods \cite{2021SVGe}.
		\label{tab:Generator_Abilites}}
	\scriptsize
	\begin{center}
		\begin{tabular}{c|c|c|c|c }
			\toprule
			Search Space  & Validity (in $\%$) & Uniqueness in ($\%$) & Novelty in ($\%$) &  Training (in GPU days)\\
			\midrule
			NAS-Bench-101 & $71.69$ & $97.92$ & $62.30$  & 0.4 \\
			NAS-Bench-201 & $99.97$ & $73.61$  & $10.03$ & 0.3\\
			NAS-Bench-301 & $42.27$  & $100$  &  $100$  & 0.9\\
			NAS-Bench-NLP & $57.95$ & $100$  & $100$  & 0.7 \\
			\bottomrule
		\end{tabular}
	\end{center}
\end{table}

\subsection{Generator Implementation Details}

In this section we present more details about the generation model SVGe from \cite{2021SVGe}.
The pseudo algorithm is described in \autoref{alg:app_decoder}. The modules $f_{\textrm{initNode}}, \\ f_{\textrm{addNode}}, f_{\textrm{addEdges}},f_{\textrm{Embedding}}$ used in this code are two-layer MLPs with ReLU activation functions. Note, in contrast to SVGe, we don't sample within the generation process, in order to allow for end-to-end learning with the prediction model for AG-Net.
\begin{algorithm*}[t]
	\SetAlgoLined
	\caption{Graph Generation}\label{alg:app_decoder}
	\label{alg:decoder}
	\KwInput{$\mathbf{z}\sim \mathcal{N}(0,1)$}
	\KwOutput{random sampled reconstructed graph $\widetilde{G}= (\widetilde{V},\widetilde{E})$} 
	initialize one-hot encoded InputNode $v_0$, with embedding $\mathbf{h}_0 \leftarrow f_{\textrm{initNode}}(\mathbf{z},f_{\textrm{Embedding}})[\textrm{InputType}])$ \\
	$V \leftarrow\{v_0\}$, $E \leftarrow \emptyset$,
	$\mathbf{h}_{{G}}$ $\leftarrow  \mathbf{{z}},  $\\
	\While{$\vert V \vert \leq \textrm{Max Number of Nodes}$}{
		$v_{t+1} \leftarrow f_{\textrm{addNode}}(\mathbf{z}, \mathbf{h}_{G})$ \\
		$ V\leftarrow V \cup \{v_{t+1}\}$  \\
		$ \mathbf{h}_{t+1} \leftarrow f_{\textrm{initNode}}(\mathbf{z}, \mathbf{h}_{G},f_{\textrm{Embedding}}(v_{t+1})])$  \\
		\For {$v_{j} \in V\setminus{v_{t+1}}$}{
			$s_{\textrm{addEdges}}(j,t+1)\leftarrow f_{\textrm{addEdges}}(\mathbf{h}_{t+1}, \mathbf{h}_t, \mathbf{h}_{G}, \mathbf{z})$  
			
			$e_{(j,t+1)} \sim \textrm{Eval}(s_{\textrm{addEdges}}(j,t+1))$ \Comment*[r]{evaluate whether to add edge} 
			\If{$e_{(j,t+1)}=1$}{
				$E \leftarrow E \cup  \{e_{(j,t+1)}= (v_j, v_{t+1})\}$  
			}
		}
		$\mathbf{h}_t \leftarrow  \textrm{concat}(\mathbf{h}_t,\mathbf{h}_{t+1})$ \\
		$G \leftarrow (V,E) $\\
		$\mathbf{h}_t \leftarrow (\mathbf{h}_t,G) $ \Comment*[r]{update node embeddings} 
		$ \mathbf{h}_{G} \leftarrow \textrm{aggregate}(\mathbf{h}_t)$ \Comment*[r]{update graph embedding } 
		$t\leftarrow t+1$ \\
	}
	$V \sim \textrm{Categorical}(V)$ \Comment*[r]{Sample node types}
	$E \sim \textrm{Ber}(E)$ \Comment*[r]{Sample edges} 
	$\widetilde{G}=(V,E)$
\end{algorithm*}

\section{Hyperparameters}\label{sec:HP}
In this section we give a detailed overview about the hyperparameter for our generative network.
We use pytorch \cite{pytorch} and pytorch geometric \cite{PYG} for all our implementations. 
\subsection{Generator}
\autoref{table:HP_generator} presents all used hyperparameters for the generation training. 
We train our generator in a ticked manner; after every $5.000$ train data, we evaluate our generator for validity ability. The used pretrained state dict for our search is then, the one, which the highest validation measurement, which is defined by randomly sample $10,000$ latent vectors $\textbf{z} \in \mathbb{R}^{32}$ and generate architectures. 
The training is the same for all different search spaces.
\begin{table}[h]
	\scriptsize
	\centering
	\caption{Hyperparameters of the generator model.}\label{table:HP_generator}
	\begin{tabular}{ c || c}
		\toprule
		
		Hyperparameter      & Default Value  \\
		\midrule
		Node Embedding      & 32        \\
		Latent Vector       & 32       \\
		MLP Node Embedding layer & 2         \\
		GNN layer           & 2       \\
		Batch Size          & 32       \\
		Optimizer            & Adam \cite{2015ADAM}     \\
		Learning Rate       &  0.0002    \\
		Betas       & (0.5, 0.999)        \\
		Ticks     & 500 \\
		Tick Size & 5,000  \\
		\bottomrule
	\end{tabular}
\end{table}

\subsection{Surrogate Model}
The overall surrogate is an MLP with ReLU activations.
\autoref{table:HP_simple_surrogate} and \autoref{table:HP_multi_surrogate} list all hyperparameters for the search experiments in the main paper for the simple performance surrogate model and the multi-objective surrogate model for the additional hardware objective.
The hyperparameters for XGB \cite{XGB} are the same as in \cite{NBSuite}. 

\begin{table}[h]
	\scriptsize
	\centering
	\caption{Hyperparameters for the performance surrogate model $f(\cdot)$}\label{table:HP_simple_surrogate}
	\begin{tabular}{c ||ccc c}
		\toprule
		Hyperparameter      & \multicolumn{4}{c}{Dataset}  \\
		\midrule
		& NB101    & NB201   & NB301 & NBNLP   \\
		\midrule
		$\alpha$              &    \multicolumn{4}{c}{0.9} \\       
		MLP Layers          &    \multicolumn{4}{c}{4}     \\
		MLP Hidden          &      56 & 84  & 176 & 559   \\
		Epochs              &    15      &  30 &  15 &  30  \\
		Optimizer           &   \multicolumn{4}{c}{Adam \cite{2015ADAM}}   \\
		LR                  & \multicolumn{4}{c}{0.001} \\     
		Betas               & \multicolumn{4}{c}{(0.5, 0.999)} \\
		weight factor       &    \multicolumn{4}{c}{10 e-3}   \\
		batch size          &   \multicolumn{4}{c}{16}  \\
		loss          &      \multicolumn{4}{c}{L2} \\
		\bottomrule
		
	\end{tabular}
\end{table}

\begin{table}[h]
	\scriptsize
	\centering
	\caption{Hyperparameters for both surrogate models $f(\cdot)$ and $g(\cdot)$ for the multi-objective search in the Hardware-Aware Benchmark}\label{table:HP_multi_surrogate}
	\begin{tabular}{c || c}
		\toprule
		
		Hyperparameter      & Hardware-Aware NASBench \\
		\midrule
		$\alpha$               &   0.95  \\  
		$\lambda$            & 0.5 \\
		MLP Layers          &       4      \\
		MLP Hidden          &      82    \\
		Epochs              &      30  \\
		Optimizer           &    Adam  \cite{2015ADAM}   \\
		LR                  &   0.002 \\     
		Betas               &  (0.5, 0.999) \\
		weight factor       &     10 e-3 \\
		penalty term         & 1000 \\
		batch size          &     16 \\
		loss                & L2\\
		\bottomrule
		
	\end{tabular}
\end{table}

\section{Latent Space Optimization Visualization}\label{sec:lso_intuition}
\begin{figure*}[h]
	\centering
	\includegraphics[width=0.8\linewidth]{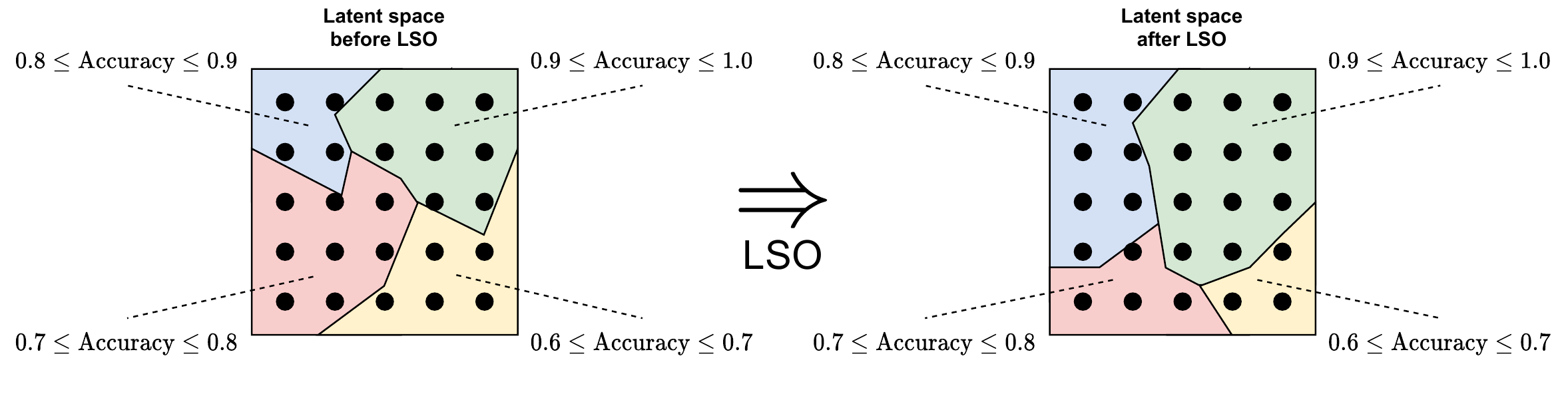}
	
	\caption{
		The latent space is reshaped in a way that promotes desired properties of generated architectures (in this example: accuracy).
		Consequently, it becomes more likely for the generator to generate architectures satisfying this property.
	}
	\label{fig:lso-intuition}
\end{figure*}

A more descriptive visualization of the latent space optimization technique used for our AG-Net neural architecture search is displayed in \autoref{fig:lso-intuition}.

\end{document}